\documentclass[pdflatex,sn-mathphys-num]{sn-jnl}%

\usepackage{graphicx}%
\usepackage{multirow}%
\usepackage{amsmath,amssymb,amsfonts}%
\usepackage{amsthm}%
\usepackage{mathrsfs}%
\usepackage[title]{appendix}%
\usepackage{xcolor}%
\usepackage{textcomp}%
\usepackage{manyfoot}%
\usepackage{booktabs}%
\usepackage{algorithm}%
\usepackage{algorithmicx}%
\usepackage{algpseudocode}%
\usepackage{listings}%
\usepackage{tabularx}
\usepackage{rotating}
\usepackage{threeparttable}
\usepackage{fvextra}
\DefineVerbatimEnvironment{Verbatim}{Verbatim}{breaklines=true, breakanywhere=true}

\theoremstyle{thmstyleone}%

\theoremstyle{thmstyletwo}%

\theoremstyle{thmstylethree}%

\begin{document}

\title[Article Title]{TrajOnco: a multi-agent framework for temporal reasoning over longitudinal EHR for multi-cancer early detection}

\author*[1,2,3]{\fnm{Sihang} \sur{Zeng}}\email{zengsh@uw.edu}

\author[1]{\fnm{Young Won} \sur{Kim}}\email{p-youngwonk@truveta.com}

\author[1]{\fnm{Wilson} \sur{Lau}}\email{chungkeil@truveta.com}

\author[1]{\fnm{Ehsan} \sur{Alipour}}\email{p-ehsana@truveta.com}

\author[2,3]{\fnm{Ruth} \sur{Etzioni}}\email{retzioni@fredhutch.org}

\author[2]{\fnm{Meliha} \sur{Yetisgen}}\email{melihay@uw.edu}

\author[1]{\fnm{Anand} \sur{Oka}}\email{anando@truveta.com}

\affil[1]{\orgname{Truveta Inc.}, \orgaddress{\city{Bellevue}, \state{WA}, \country{United States}}}

\affil[2]{\orgname{University of Washington}, \orgaddress{\city{Seattle}, \state{WA}, \country{United States}}}

\affil[3]{\orgname{Fred Hutch Cancer Center}, \orgaddress{\city{Seattle}, \state{WA}, \country{United States}}}

\abstract{Accurate estimation of cancer risk from longitudinal electronic health records (EHRs) could support earlier detection and improved care, but modeling such complex patient trajectories remains challenging. We present TrajOnco, a training-free, multi-agent large language model (LLM) framework designed for scalable multi-cancer early detection. Using a chain-of-agents architecture with long-term memory, TrajOnco performs temporal reasoning over sequential clinical events to generate patient-level summaries, evidence-linked rationales, and predicted risk scores. We evaluated TrajOnco on de-identified Truveta EHR data across 15 cancer types using matched case-control cohorts, predicting risk of cancer diagnosis at 1 year. In zero-shot evaluation, TrajOnco achieved AUROCs of 0.64-0.80, performing comparably to supervised machine learning in a lung cancer benchmark while demonstrating better temporal reasoning than single-agent LLMs. The multi-agent design also enabled effective temporal reasoning with smaller-capacity models such as GPT-4.1-mini. The fidelity of TrajOnco's output was validated through human evaluation. Furthermore, TrajOnco’s interpretable reasoning outputs can be aggregated to reveal population-level risk patterns that align with established clinical knowledge. These findings highlight the potential of multi-agent LLMs to execute interpretable temporal reasoning over longitudinal EHRs, advancing both scalable multi-cancer early detection and clinical insight generation.}

\keywords{Multi-Cancer Early Detection, Large Language Models, Multi-Agent System, Patient Trajectory}

\maketitle

\section{Introduction}
\label{sec:Introduction}

Early detection is an aspirational approach for significantly reducing the burden of cancer morbidity and mortality \cite{whitaker2020earlier, crosbyEarlyDetectionCancer2022}.
There is an emerging interest in harnessing large-scale longitudinal electronic health record (EHR) data to model multi-cancer risk \cite{jung2024multi, zhao2025development, shmatkoLearningNaturalHistory2025}. Modeling this risk generally serves two primary purposes: tailoring screening and surveillance for high-risk individuals, or advancing a future cancer diagnosis for those already exhibiting subtle clinical signals. EHR data contain detailed records of patient trajectories capturing these clinical signals, such as diagnoses, laboratory tests, and medications over time. For example, gradually worsening anemia and repeated imaging for unexplained abdominal pain may precede the diagnosis of certain cancers \cite{holtedahlAbdominalSymptomsCancer2018, mondocImpactAnemiaSurvival}. Identifying these evolving clinical patterns offers a potential modality built directly into routine clinical care to advance the time of clinical diagnosis.

Depending on the time horizon before a cancer diagnosis and the true disease status of the patient, EHR-based prediction problems can generally be categorized into three distinct tasks. First, \textit{near-term diagnosis prediction} addresses the very short-term horizon (weeks to months) where a patient is already exhibiting obvious clinical symptoms, aiming to predict an imminent diagnosis \cite{liCancerLLMLargeLanguage2026}. Second, \textit{early detection} (or advanced clinical diagnosis) utilizes a moderate time horizon (1-3 years) to predict short-term cancer risk based on early, subtle, or evolving clinical signals before full symptomatic disease is recognized. Finally, \textit{long-term risk prediction} targets a longer time horizon (5+ years) to predict future cancer risk in patients who currently do not have the disease, guiding tailored surveillance. These settings represent increasing levels of difficulty as signals become weaker and more temporally diffuse. Specifically for early detection, capturing these subtle, accumulating signals across a patient's trajectory requires robust temporal reasoning over longitudinal EHR data \cite{whitaker2020earlier, crosbyEarlyDetectionCancer2022}.

Prior studies detecting early cancer from EHRs have predominantly relied on data from large national healthcare systems \cite{jung2024multi, shmatkoLearningNaturalHistory2025}. Multi-cancer early detection with EHRs across diverse, multi-system environments remains challenging. Existing approaches typically rely on supervised machine learning (ML) models that require extensive feature engineering and training data \cite{parkScalableEarlyCancer2026, zhao2025development, jung2024multi}. In addition, their outputs and interpretations, such as hazard ratios or SHAP values \cite{lundbergUnifiedApproachInterpreting2017a}, offer limited insight into the complex temporal patterns within patient histories that jointly shape the evolution of cancer risk.

Large language models (LLMs) have emerged as a flexible and generalizable paradigm, offering evidence-driven reasoning with improved interpretability. In oncology, prior applications have focused on short-context tasks such as cancer progression prediction, phenotype extraction, and diagnosis generation from individual notes or reports \cite{liCancerLLMLargeLanguage2026, zhuLargeLanguageModel2025, lauPredictingEarlyOnsetColorectal2025}. However, extending LLMs to early detection requires modeling temporal patterns and interactions among diverse clinical factors over long patient trajectories which may span several years. Conventional single-agent LLMs struggle with such long-context reasoning due to limitations such as the “lost-in-the-middle” effect \cite{liu-etal-2024-lost}, which may lead to the omission of evolving EHR signals. Agent-based LLM systems have recently been proposed to decompose complex reasoning into coordinated steps across multiple agents \cite{liCAREADMultiagentLarge2025}, offering a promising approach for temporal reasoning over long and heterogeneous EHR data.

In this study, we introduce TrajOnco, a training-free multi-agent framework for temporal reasoning over longitudinal EHR data in multi-cancer early detection. 
We define multi-cancer early detection as prediction of short-term risk across multiple distinct cancer types within a shared modeling framework. TrajOnco addresses this setting in a zero-shot manner using the same base model and architecture across cancers, with only minimal prompt adaptation to specify the target cancer type.
The TrajOnco framework processes patient trajectories sequentially using a chain-of-agents (CoA) architecture \cite{zhang2024chain}, augmented with a long-term memory (LTM) module to retain a global context with critical cancer-related events.

Using de-identified structured EHR data encompassing multiple healthcare systems from Truveta, we evaluated TrajOnco to predict cancer risk across 15 cancer types. To focus on early detection and avoid overlap with near-term diagnosis prediction, we imposed a 1-year gap between the time of prediction and the diagnosis or index date (Fig. \ref{fig:architecture}b), leveraging all available EHR data prior to the time of prediction. In zero-shot settings, TrajOnco achieved AUROCs ranging from 0.64 to 0.80 across cancers and performed comparably to supervised ML models in a lung cancer benchmark. Notably, its multi-agent design enables effective temporal reasoning without relying on very large models. 
In addition to individual-level predictions, TrajOnco identifies key clinical events associated with cancer risk, informing our understanding of the most common factors that portend a future cancer diagnosis.

\section{Results}
\label{sec:Results}

\subsection{Overview of TrajOnco}
\begin{figure}[h]
    \centering
    \includegraphics[width=\linewidth]{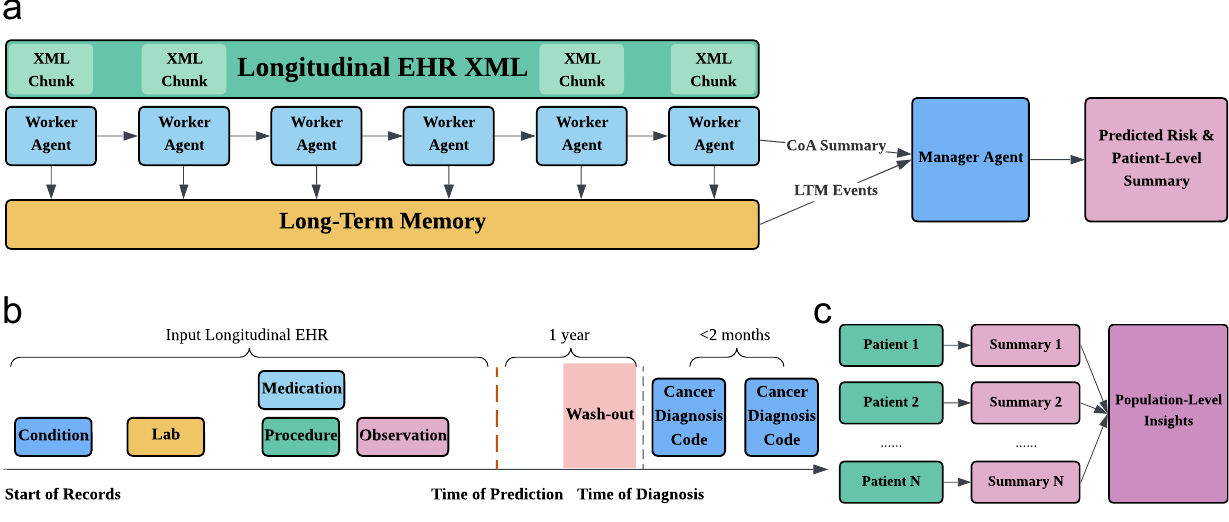}
    \caption{\textbf{TrajOnco framework and study design.} \textbf{a.} TrajOnco uses sequential worker agents, long-term memory, and a manager agent to process longitudinal EHR and produce predicted risk with a patient-level summary. \textbf{b.} In this early detection task, EHR data from the start of records to the prediction time were used to predict risk of cancer diagnosis at 1 year. Two cancer diagnosis codes within 2 months were used to define time of cancer diagnosis. \textbf{c.} Patient-level summaries can be aggregated to generate population-level insights.}
    \label{fig:architecture}
\end{figure}

TrajOnco is a training-free multi-agent framework designed to model long and heterogeneous longitudinal EHR data for multi-cancer early detection (Fig. \ref{fig:architecture}a). It builds on a chain-of-agents (CoA) architecture \cite{zhang2024chain}, where patient trajectories are converted into a unified, time-ordered XML representation and partitioned into temporally coherent chunks, preserving rich information in an LLM-friendly format while avoiding costly task-specific feature engineering and data cleaning. Sequential worker agents process these chunks, extracting salient cancer-related events and updating a cumulative summary with risk assessments (low, moderate, or high) and supporting rationale. This decomposition reduces per-agent context length, mitigates the “lost-in-the-middle” problem \cite{liu-etal-2024-lost}, and enables progressive temporal reasoning over long patient histories.

To prevent catastrophic forgetting of early but clinically important events, TrajOnco maintains a structured long-term memory (LTM) that stores deduplicated, time-stamped risk factors and key events identified by worker agents, providing a global context across the patient trajectory. A manager agent then synthesizes the final cumulative unstructured worker summary together with the structured LTM to generate a cancer-specific risk estimate (integer between 1 and 10), an evidence-based narrative justification, and final identified cancer-related events. This design separates local signal extraction from global reasoning, enabling the modeling of long-range temporal dependencies without additional training. The framework generalizes across cancer types with minimal prompt modifications (e.g., substituting lung cancer with liver cancer), while its explicit memory and summaries support interpretable patient-level evidence and aggregation of population-level insights (Fig. \ref{fig:architecture}c).

\subsection{Zero-shot multi-cancer early detection}

Truveta data comprise de-identified EHRs from over 30 US health systems, covering more than 120 million patients across 900 hospitals and 20,000 clinics. Using these data, we identified cohorts for 15 cancer types, including liver, lung, prostate, breast, bladder, cervical, esophageal, endometrial, gastric, ovarian, pancreatic, colorectal cancers, leukemia, multiple myeloma, and lymphoma, based on clinician-curated diagnosis codes (Appendix \ref{sec:data_statistics}). These represent major solid and hematologic malignancies commonly studied in early detection research \cite{jung2024multi, parkScalableEarlyCancer2026}. Incident cancer was defined based on two diagnosis codes within a 2-month interval after a 6-month cancer-free washout period. The date of the first diagnosis code in the first qualifying pair was defined as the cancer diagnosis date \cite{haugNewonsetCancerCases2022, setoguchiAgreementDiagnosisIts2007}. The 6-month cancer-free washout period was used to exclude prevalent cases \cite{setoguchiAgreementDiagnosisIts2007}. We used structured EHR data from the start of records to 1 year prior to diagnosis (or a randomly assigned index date for controls) to predict risk of cancer diagnosis at 1 year (Fig. \ref{fig:architecture}b). 

To evaluate TrajOnco’s performance across cancer types, we conducted three analyses: (i) cancer-specific case-control prediction for each cancer (Section \ref{sec:per-cancer}), (ii) benchmarking on a lung cancer cohort against established machine learning models (Section \ref{sec:benchmark}), and (iii) sensitivity analyses on efficiency and scalability (Section \ref{sec:tradeoff}). Unless otherwise specified, GPT-4.1-mini was used as the base model for both worker and manager agents due to its favorable balance of performance, latency, and cost. The context length of each XML chunk was set as 16k tokens. We also evaluated a composite any-cancer prediction task following prior work \cite{zhao2025development} (Appendix \ref{sec:any-cancer}).

\subsubsection{Cancer-specific prediction across 15 cancers}
\label{sec:per-cancer}

For each cancer type, we constructed a case-control test set comprising 500 randomly sampled cases and 500 controls. This moderate sample size was determined to balance the variance in performance estimates and the experimental costs. Controls were randomly sampled and matched 1:1 to cases by sex and 10-year age group to reduce demographic confounding. For TrajOnco, worker and manager agent prompts were kept consistent across cancer types, with only the cancer type modified, ensuring comparable model behavior (Appendix \ref{sec:prompts}).

\begin{figure}[h]
    \centering
    \includegraphics[width=\linewidth]{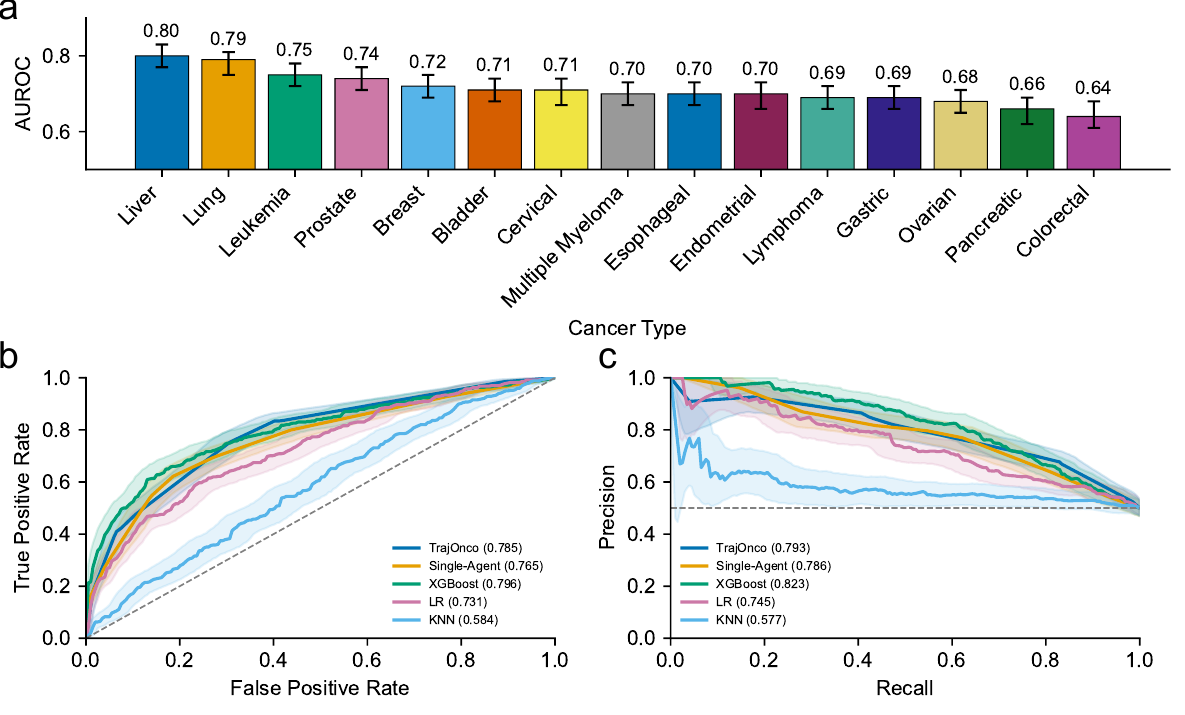}
    \caption{\textbf{Zero-shot prediction performance of TrajOnco.} \textbf{a.} Area under receiver operating characteristic (AUROC) of cancer-specific prediction on each cancer type; \textbf{b.} ROC curves and \textbf{c.} precision-recall curves for different methods on the lung cancer benchmark.}
    \label{fig:performance}
\end{figure}

Model discrimination varied across the 15 cancer types (AUROC 0.64-0.80; Fig. \ref{fig:performance}a). Performance was highest for liver and lung cancers and lowest for colorectal cancer, while most other solid and hematologic malignancies demonstrated comparable discrimination with AUROCs above 0.70. This indicates that the framework generalizes across diverse cancer types, albeit with cancer-specific variations in predictive accuracy.

\subsubsection{Benchmarking on lung cancer cohort}
\label{sec:benchmark}

To benchmark zero-shot performance, we compared TrajOnco with logistic regression (LR), k-nearest neighbors (KNN), XGBoost \cite{chenXGBoostScalableTree2016}, and a single-agent baseline on an age- and sex-matched lung cancer cohort of 500 random cases and 500 controls. For supervised models, we derived summary feature vectors from longitudinal EHR data, retaining features with at least 1\% frequency (1,253 features) following prior work \cite{liEarlyDetectionNonsmall2025}. The single-agent baseline used the same base LLM, the same XML input representation, and the same final risk output format as TrajOnco, but processed the entire available input in one pass. Supervised models were trained and validated on a separate dataset of 170,464 cases and 166,903 controls using an 8:2 split, while LLM-based methods remained zero-shot prediction.

On this benchmarking cohort, TrajOnco achieved an AUROC of 0.785, outperforming the single-agent baseline (AUROC = 0.765) and substantially exceeding the performance of logistic regression and KNN, while performing slightly below XGBoost (AUROC = 0.796) (Fig. \ref{fig:performance}b,c). These results indicate that TrajOnco, despite not being trained specifically for cancer prediction, achieves performance comparable to a trained gradient boosting model and markedly surpasses conventional linear and distance-based approaches for lung cancer early detection.

\begin{figure}[h]
    \centering
    \includegraphics[width=\linewidth]{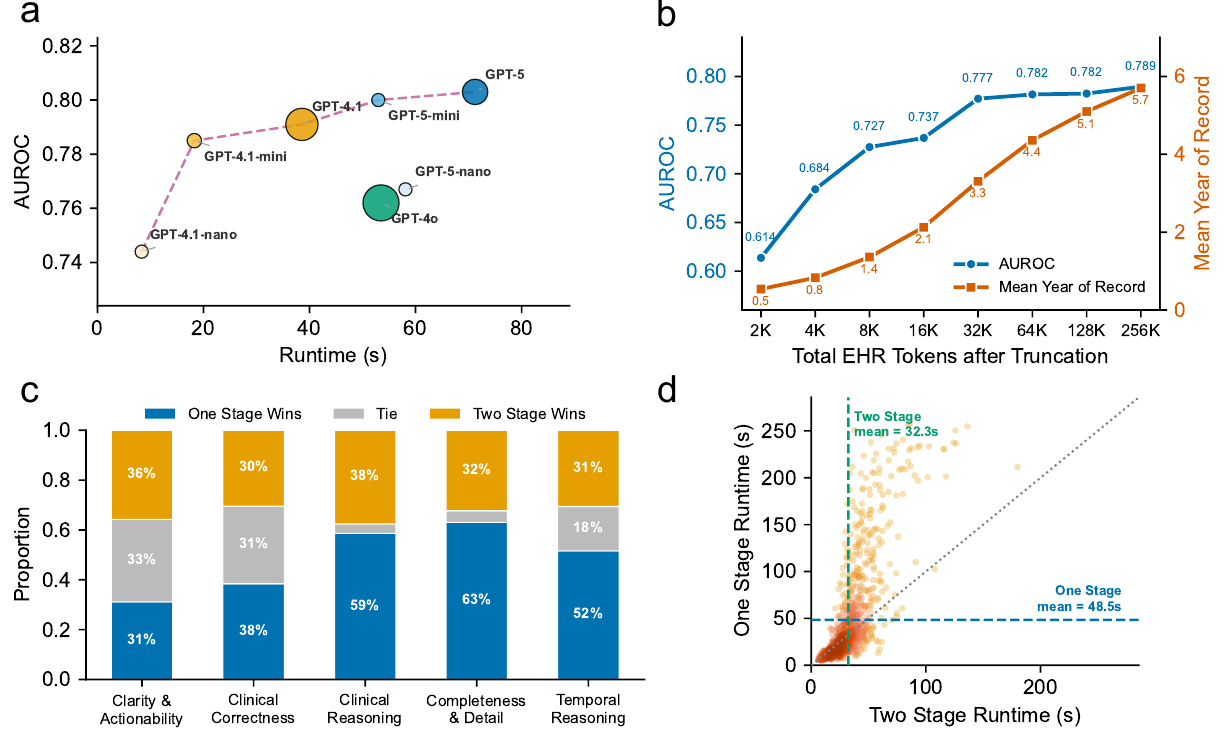}
    \caption{\textbf{Sensitivity analyses on the scalability and efficiency of TrajOnco.} \textbf{a.} AUROC and median runtime of TrajOnco on the lung cancer benchmark using different base models. The radius of each point is proportional to the API price of the input token of the model. The pink dashed line represents the Pareto front. Among these base models, GPT-4.1-mini achieved a balance between performance, latency, and cost. \textbf{b.} Improved predictive performance when expanding the maximum length of left-truncated patient trajectory. \textbf{c.} LLM-as-a-judge comparison between one-stage TrajOnco and two-stage TrajOnco. The one-stage model had better clinical reasoning, completeness of detail, and temporal reasoning than the two-stage model. \textbf{d.} The two-stage model achieved lower latency especially for very long EHR sequences.}
    \label{fig:sensitivity}
\end{figure}

\subsubsection{Sensitivity analyses}
\label{sec:tradeoff}

To evaluate the scalability and efficiency of TrajOnco, we conducted sensitivity analyses examining (1) base model selection, (2) patient trajectory length, and (3) context detailedness.

\paragraph{Base model selection}

We replaced the default base model (GPT-4.1-mini) with alternative GPT models, including GPT-4o, GPT-4.1-nano, GPT-4.1, GPT-5-nano, GPT-5-mini, and GPT-5. For reasoning models, we used a medium reasoning effort setting. All models were evaluated within the identical TrajOnco multi-agent framework on the lung cancer benchmark.

We observed that more advanced reasoning models (e.g., GPT-5) did not substantially outperform smaller models such as GPT-4.1-mini (Fig. \ref{fig:sensitivity}a). We hypothesize that this is because the TrajOnco architecture decomposes complex EHR understanding into smaller, well-defined subtasks with structured inter-agent communication and summarization. This design reduces the need for complex global reasoning within a single model invocation, thereby diminishing the performance advantage typically associated with larger reasoning models on complex tasks.

Within each model family, larger models generally achieved improved predictive performance, but at the expense of higher computational cost and increased latency (Fig. \ref{fig:sensitivity}a). Overall, GPT-4.1-mini provided the most favorable balance between runtime, cost, and predictive performance, and was therefore selected as the default base model.

\paragraph{Patient trajectory length}

To assess the impact of patient trajectory length, i.e., how far back in a patient’s history TrajOnco uses for prediction, we conducted experiments on the lung cancer benchmark using left-truncated EHRs with varying maximum token limits, simulating scenarios where only partial historical data are available. Specifically, we truncated the input EHR XML to retain only the most recent 2k to 256k tokens before prediction.

Increasing the trajectory length from 2k to 256k tokens extended the average history from 0.5 to 5.7 years and improved AUROC from 0.614 to 0.789 on the same benchmark (Fig. \ref{fig:sensitivity}b). These results indicate that longer patient trajectories consistently improve predictive performance, suggesting that richer longitudinal clinical information provides stronger signals for cancer risk assessment. Overall, the findings demonstrate that TrajOnco benefits from extended patient records when available, supporting its scalability and applicability to comprehensive real-world EHR data.

\paragraph{Context detailedness}

For patients with very long EHRs, the number of sequential worker agents increases proportionally, resulting in high latency due to the inherently sequential design of the framework. To mitigate this limitation, we evaluated a two-stage architecture (Fig. \ref{fig:two_stage}). In this variant, each EHR chunk was first processed independently by a preprocessor agent in parallel. The preprocessor extracted informative events, which were then concatenated into a shorter, structured XML representation for downstream consumption by the TrajOnco pipeline. This design reduces the number of sequential worker agents and converts part of the workflow from sequential to parallel execution.

The two-stage approach reduced average latency by 25\% relative to the one-stage model (32.3 vs. 48.5 s), with larger gains in patients with very long EHRs (Fig. \ref{fig:sensitivity}d). Theoretical complexity analysis was provided in Appendix \ref{sec:two-stage}. Predictive performance was largely preserved (AUROC, 0.78 vs. 0.79). However, compared with the original TrajOnco framework, the two-stage model produced less detailed summaries and weaker clinical reasoning (Fig. \ref{fig:sensitivity}c), likely because parallel preprocessors operated without global context, resulting in less complete information extraction and a more abstract XML representation of the EHR. These findings indicate that parallel preprocessing in the two-stage model improves efficiency for very long EHRs, but may reduce summary fidelity and reasoning quality, highlighting a trade-off between latency and interpretability.

\subsection{TrajOnco generates clinically meaningful patient-level summaries}

Beyond producing a final risk score, TrajOnco's sequential worker agents maintain an evolving summary over time, while the manager agent generates a final structured patient-level summary for each individual. These summaries provide a condensed view of the patient trajectory, enhancing interpretability. We evaluated the quality and clinical relevance of these summaries through: (i) comparison with a single-agent baseline using an LLM-as-a-judge framework \cite{zhengJudgingLLMasaJudgeMTBench2023} (Section \ref{sec:llm_judge}), (ii) human evaluation assessing the fidelity of the summary (Section \ref{sec:human_eval}), and (iii) a case study illustrating how the worker agents dynamically aggregate clinical events and update the predicted risk (Section \ref{sec:case_study}).

\begin{figure}[h]
    \centering
    \includegraphics[width=\linewidth]{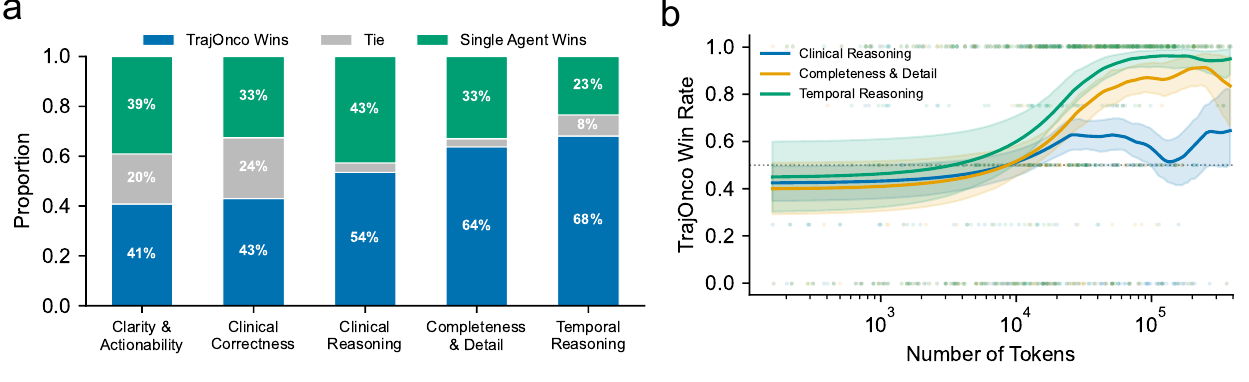}
\caption{\textbf{LLM-as-a-judge evaluation of TrajOnco versus a single-agent baseline.} \textbf{a,} Pairwise comparison across five dimensions using GPT-5 with high reasoning effort as the judge. Bars show the proportions of TrajOnco wins (blue), ties (gray), and single-agent wins (green). TrajOnco outperforms the baseline across all dimensions, with the largest advantage in Temporal Reasoning (68\% wins vs. 23\% losses). \textbf{b,} TrajOnco win rate by EHR length (input tokens; log scale) for selected dimensions. Solid lines indicate smoothed win rates and shaded areas indicate confidence intervals. Each scattered point represents the average win rate over two runs with different candidate orders. For each run, ``win" assigns a score of 1.0, ``tie" assigns a score of 0.5, and ``loss" assigns a score of 0.0.}
    \label{fig:judge}
\end{figure}

\subsubsection{Comparison with a single-agent baseline using LLM-as-a-judge evaluation}
\label{sec:llm_judge}

TrajOnco was designed to enhance temporal reasoning over longitudinal EHR through its CoA and LTM architecture. To assess whether this design improves summary quality, we conducted a pairwise LLM-as-a-judge evaluation \cite{zhengJudgingLLMasaJudgeMTBench2023} comparing TrajOnco with a single-agent LLM baseline on the lung cancer case cohort. Evaluations were performed using GPT-5 configured with high reasoning effort. We defined five prespecified evaluation dimensions: (i) Clarity \& Actionability, (ii) Clinical Correctness, (iii) Clinical Reasoning, (iv) Completeness \& Detail, and (v) Temporal Reasoning. To mitigate the position bias from the order of two candidates \cite{shiJudgingJudgesSystematic2025}, we report the average win/tie/loss rates over two runs with different candidate orders. 

Across all five dimensions, TrajOnco achieved more wins than losses compared to the single-agent model (Fig. \ref{fig:judge}a), with the largest advantage in temporal reasoning (68\% wins vs. 23\% losses). It also outperformed in completeness and detail (64\% vs. 33\%) and clinical reasoning (54\% vs. 43\%), indicating more comprehensive and better-integrated summaries. Notably, this advantage increased with longer EHRs (Fig. \ref{fig:judge}b), suggesting that the multi-agent design and memory mechanisms are particularly effective for long and complex patient trajectories.

\subsubsection{Human evaluation}
\label{sec:human_eval}
To assess the fidelity of TrajOnco-generated summaries, we conducted a human evaluation with two independent annotators (a PhD candidate in biomedical informatics and a medical doctor). Annotators assessed whether each identified cancer-related event accurately reflected the corresponding EHR evidence, labeling events as \textit{correct} or \textit{inaccurate}. We randomly sampled 30 patients from 15 cancer case-control test sets, yielding 113 events. Error modes were then summarized from the inaccurate events. Disagreements between annotators were resolved through discussion.

Cohen’s kappa was 0.72, indicating substantial inter-annotator agreement. Of the 113 evaluated events, 102 (90.27\%) were annotated as correct. The remaining errors were primarily attributable to temporal inaccuracies (n = 5) and interpretive inaccuracies (n = 6). Temporal errors typically occurred when TrajOnco identified an event as ``new" even though it had been documented previously. Interpretive errors were mainly due to minor hallucinations or overly assertive inferences, such as adding unsupported qualifiers (e.g., referring to a ``\textit{fasting} glucose test" when fasting status was not documented in the original EHR) or failing to preserve clinical uncertainty. In some cases, the system described findings such as ``suggestive of disease progression" or ``significantly elevating liver cancer risk" although the available evidence could also have been consistent with alternative explanations, such as nonspecific abnormalities or a benign lesion. Overall, these inaccuracies were infrequent and did not materially compromise the clinical validity or usefulness of the patient summaries.

\begin{figure}[h]
    \centering
    \includegraphics[width=\linewidth]{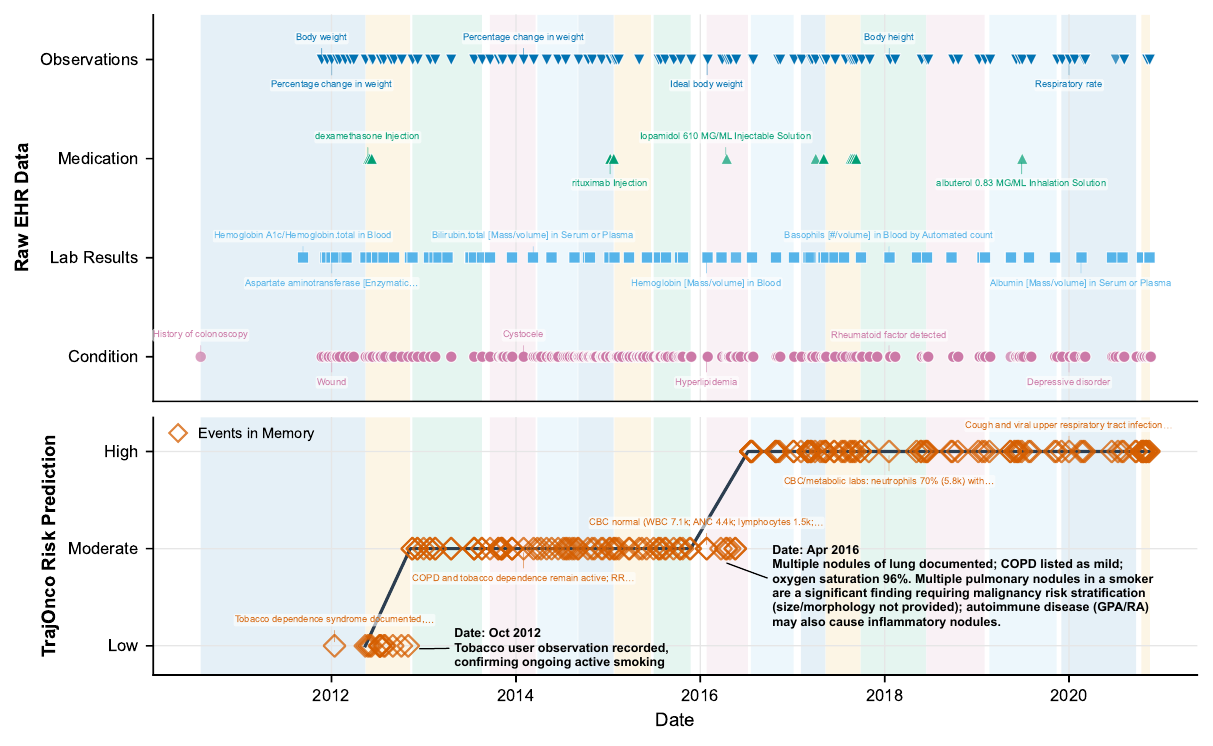}
    \caption{\textbf{Dynamic short-term risk prediction and event aggregation by TrajOnco worker agents.} The top panel displays the patient's raw EHR data, encompassing observations, medications, lab results, and conditions from 2012 to 2020. Different temporal chunks processed sequentially by the worker agents are distinguished by alternating background colors. The bottom panel shows TrajOnco's continuously updated short-term risk prediction corresponding to the timeline. Key clinical insights synthesized by the agents and stored as events in LTM are explicitly marked with diamond symbols along the risk trajectory. Two critical updates are highlighted: (1) In October 2012, an observation confirming ongoing active smoking shifts the baseline risk. (2) In April 2016, the discovery of multiple pulmonary nodules and mild COPD triggers a risk escalation to high, with the system's memory specifically noting the need for malignancy risk stratification in a smoker, while remaining robust to alternative inflammatory etiologies such as autoimmune disease.}
    \label{fig:case_study}
\end{figure}

\subsubsection{Case study}
\label{sec:case_study}
To illustrate the dynamic nature of TrajOnco's risk assessment, we present a longitudinal case study tracking a single patient's clinical trajectory from 2012 to 2020 (Fig. \ref{fig:case_study}). As worker agents process temporal EHR chunks, they extract key events into LTM and continuously update risk. For example, active smoking in 2012 was recognized and increased the risk to moderate, while the emergence of multiple lung nodules in 2016, combined with prior smoking history and COPD, escalated the risk to high. This demonstrates how TrajOnco's sequential agents conduct clinical reasoning by synthesizing historical context with emerging events to dynamically update patient-level risk.

\subsection{TrajOnco generates population-level insights in cancer risk}

Each patient-level summary contains the cancer-related events identified by TrajOnco, enabling aggregation into population-level insights for the cancer. In this section, (i) we use lung cancer as a case study to characterize the thematic EHR signals underlying TrajOnco’s risk estimates for this cancer type (Section \ref{sec:lung_population}), and (ii) we summarize the top five themes identified across all 15 cancer types together with the cross-cancer patterns derived from these summaries (Section \ref{sec:cross_cancer_themes}). In the Appendix \ref{sec:sankey}, we further analyze shared temporal patterns in predicted risk trajectories and the reasons for risk transitions derived from worker agents' outputs. These analyses illustrate TrajOnco’s utility not only as a predictive model, but also as a framework for generating interpretable population-level insights from longitudinal EHR data.

\begin{figure}[h!]
    \centering
    \includegraphics[width=\linewidth]{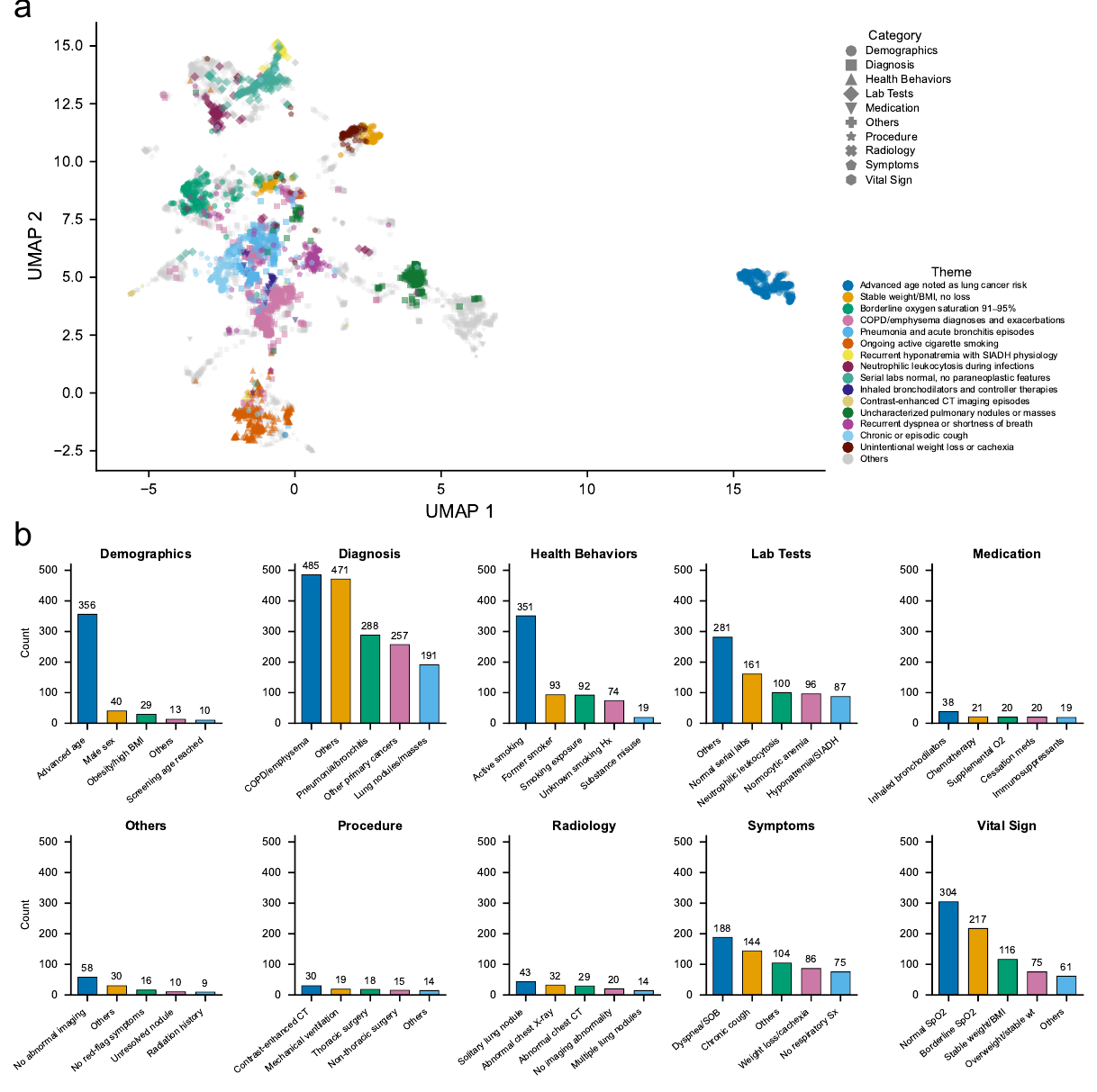}
    \caption{\textbf{Thematic patterns of lung cancer-related events identified by TrajOnco among case patients.} \textbf{a.} UMAP projection of cancer-related events output by TrajOnco, demonstrating coherent clustering of events into clinically interpretable themes across demographic factors, comorbid diagnoses, health behaviors, laboratory findings, medications, procedures, radiology, symptoms, and vital signs. \textbf{b.} Distribution of the most frequently identified events within each category, highlighting dominant population-level signals such as smoking exposure. Note that the counts were the mentions of all identified events within each category, which can be more than one for each patient.}
    \label{fig:lung_cancer_insight}
\end{figure}

\subsubsection{Thematic EHR signals underlying lung cancer risk estimation}
\label{sec:lung_population}
We use lung cancer as an example to illustrate how the cancer-related events in the final summaries can be aggregated for population-level interpretation.
Specifically, we applied a topic modeling approach similar to TopicGPT \cite{phamTopicGPTPromptbasedTopic2024} to uncover common themes within these cancer-related events. In lung cancer, the events identified by TrajOnco form distinct, clinically interpretable clusters (Fig. \ref{fig:lung_cancer_insight}a), with the dominant themes (Fig. \ref{fig:lung_cancer_insight}b) reflecting established risk factors, comorbidities, and early symptoms across various event categories, including advanced age \cite{liAssociationAgeLung2025}, smoking exposure, COPD \cite{walserSmokingLungCancer2008}, respiratory symptoms like shortness of breath and chronic cough \cite{pradoSymptomsSignsLung2023}, lung nodules \cite{loverdosLungNodulesComprehensive2019}, and laboratory abnormalities like anemia \cite{pirkerAnemiaLungCancer2003}. TrajOnco also captured signals associated with benign or stable trajectories, such as normal imaging and serially normal laboratory findings, indicating that its predictions integrate both positive and negative evidence. These patterns not only illustrate that the aggregation of TrajOnco's output summaries recovers known clinical associations, but also show how TrajOnco integrates these multidimensional clinical signals into interpretable estimates of lung cancer risk.

\begin{figure}[h]
    \centering
    \includegraphics[width=\linewidth]{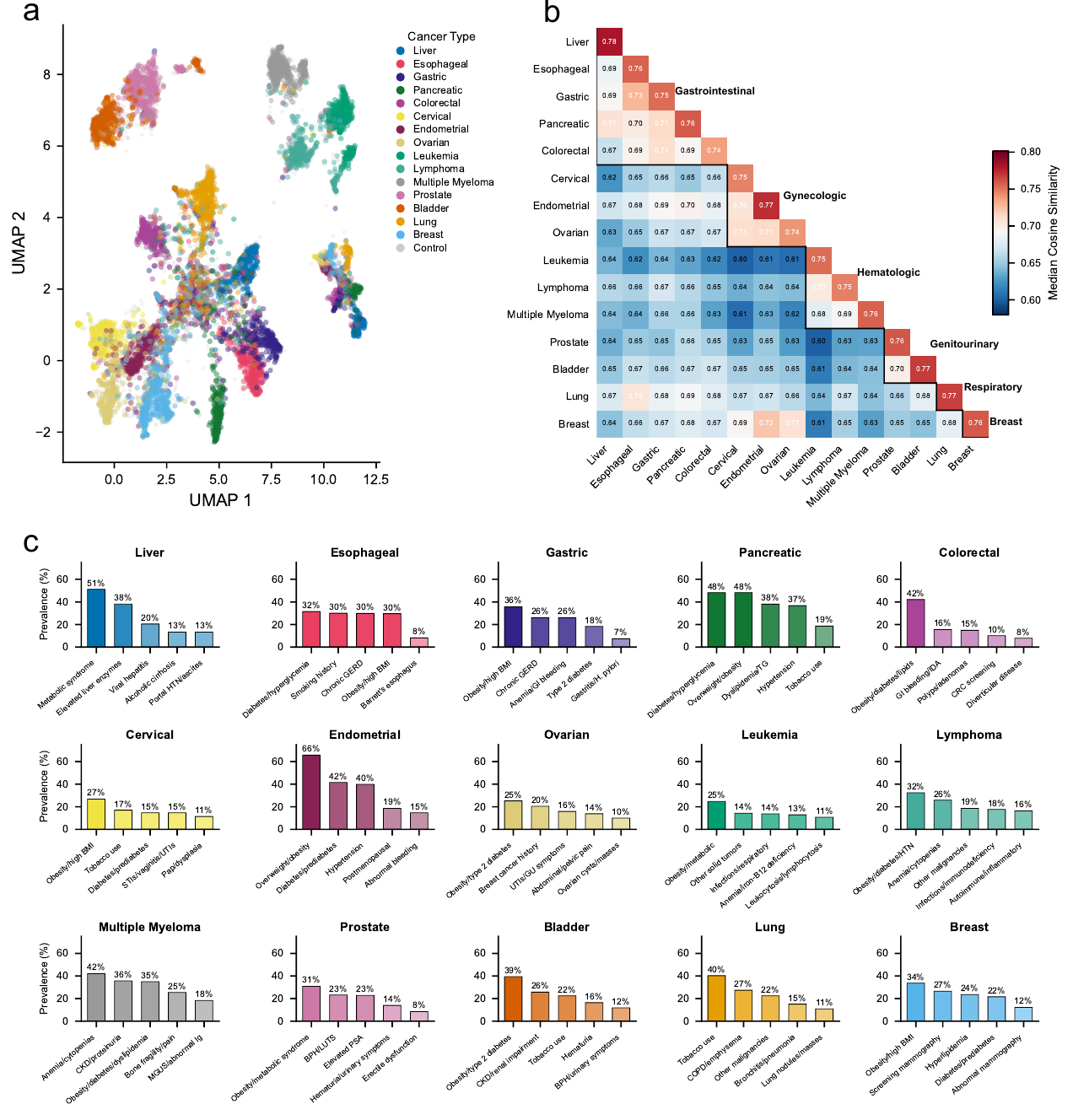}
    \caption{\textbf{Cancer-specific themes and relationships derived from patient summaries.} \textbf{a.} UMAP of patient summaries colored by cancer type, showing that patients with the same cancer tend to cluster together. \textbf{b.} Heatmap of median cosine similarity between patients from pairwise cancer types, highlighting higher within-cancer similarity and relationships among related cancers. \textbf{c.} Top prevalent themes identified for each cancer type from patient summaries.}
    \label{fig:all_cancer_events}
\end{figure}

\subsubsection{Cross-cancer relationships}
\label{sec:cross_cancer_themes}

Extending the topic modeling analysis across all 15 cancer types showed that TrajOnco can generate interpretable, data-driven insights across diseases (Fig. \ref{fig:all_cancer_events}). Using each patient’s full set of identified cancer-related events as an individual summary, the summary-level UMAP analysis \cite{mcinnesUMAPUniformManifold2020} revealed that patients with the same cancer type tend to occupy nearby regions of the latent space, indicating that TrajOnco captures meaningful within-cancer signal structure rather than diffuse, nonspecific comorbidity patterns (Fig. \ref{fig:all_cancer_events}a). Consistent with this, the similarity matrix showed greater thematic resemblance within major cancer groupings, including gastrointestinal, hematologic, and genitourinary cancers, while still preserving disease-specific profiles (Fig. \ref{fig:all_cancer_events}b). The prevalence analysis further clarified the dominant signals underlying each cancer risk and cross-cancer relationships, consistent with clinical knowledge. For instance, colorectal cancer was characterized by metabolic factors \cite{yuDiabetesColorectalCancer2022, mirandaObesityColorectalCancer2024} and iron-deficiency anemia \cite{chardaliasIronDeficiencyAnemia2023} (Fig. \ref{fig:all_cancer_events}c). From these prevalent themes, common themes among cancer types can be derived. Examples of these derived cross-cancer relationships include that gastrointestinal cancers were linked by metabolic and nutrition-related themes \cite{cencioniGastrointestinalCancerPatient2022}, hematologic malignancies were linked by cytopenia- and anemia-related symptoms \cite{johnsonHematologicManifestationsMalignancy1989a}, and lung and bladder cancer shared smoking-associated signals \cite{vainribUrologicalImplicationsConcurrent2007}. Obesity-related and broader metabolic themes recurred across multiple cancers. These results highlight TrajOnco’s utility not only for prediction, but also for generating population-level insight into shared and cancer-specific themes from longitudinal EHR data.

\section{Discussion}
\label{sec:Discussion}

In this study, we developed TrajOnco, a training-free multi-agent framework for 1-year multi-cancer early detection from longitudinal structured EHR data. Across 15 cancer types, TrajOnco showed variable discrimination in a zero-shot setting, with AUROCs ranging from 0.64 to 0.80, and the majority of them exceeding 0.70. It achieved performance on the lung cancer benchmark comparable to that of a supervised XGBoost model, despite requiring no task-specific training. Beyond risk estimation, the framework generated patient-level summaries linked to temporally ordered clinical evidence and recovered clinically coherent population-level themes across cancers, suggesting TrajOnco's potential in both prediction and interpretation from routine EHR data.

The main technical contribution of TrajOnco is a zero-shot, multi-agent framework that combines temporal decomposition with explicit memory to enable interpretable and generalizable modeling of long patient trajectories. By converting multimodal EHR data into a unified, LLM-friendly XML representation and organizing it into temporally coherent chunks, TrajOnco avoids extensive feature engineering, which is costly and difficult to generalize. Its CoA design, coupled with LTM, separates local signal extraction from global synthesis, allowing the model to capture complex temporal dependencies that are difficult for single-agent LLMs due to long-context limitations such as the ``lost-in-the-middle" effect \cite{liu-etal-2024-lost}.

Compared to conventional EHR ML models that require task-specific training \cite{liEarlyDetectionNonsmall2025, pmlr-v298-zeng25a}, EHR foundation models with limited interpretability \cite{zhangExploringScalingLaws2025a, rencZeroShotHealth2024, shmatkoLearningNaturalHistory2025}, and prior single-agent LLM approaches that struggle with long-context reasoning \cite{cuiTIMERTemporalInstruction2025a, kruseZeroshotLargeLanguage2025, wornow2024context}, TrajOnco explicitly structures temporal reasoning and embeds interpretability directly into the inference process through evidence-linked summaries and extracted clinical events. Its zero-shot formulation further enables generalization across cancer types through minimal prompt adaptation. With similar prompt adaptation, TrajOnco has the potential to be extended to other clinical prediction tasks.

From a clinical perspective, TrajOnco lies in the area of multi-cancer early detection, where longitudinal history, risk factors, early symptoms, and physiologic changes may jointly indicate short-term cancer risk. This setting is clinically relevant because it could support earlier diagnosis and more timely screening \cite{crosbyEarlyDetectionCancer2022}. 
Prior EHR-based approaches have shown promise in multi-cancer early detection \cite{jung2024multi, parkScalableEarlyCancer2026}, but have often required supervised training and offered limited interpretability.  
In contrast, TrajOnco harnesses LLMs to demonstrate consistent \textit{zero-shot} performance on Truveta data spanning more than 30 US healthcare systems and produces rich thematic interpretation from the summaries consistent with clinical knowledge. 

TrajOnco demonstrated variable performance across the evaluated cancer types, with the strongest performance observed in liver cancer and more challenging results in colorectal cancer, aligning with prior studies \cite{zhao2025development, jung2024multi}. It remains unclear exactly why this variability exists. It is possible that some cancers inherently lack detectable clinical signals in structured EHR data more than 1 year prior to diagnosis, often presenting instead with nonspecific symptoms closer to diagnosis \cite{duanColorectalCancerOverview2022}. To analyze whether this was the case, we performed sensitivity analyses using time gaps of 0.5, 1, 2, 3, 4, and 5 years (Appendix \ref{sec:sensitivity_time_gap}). Performance generally improved as the time gap decreased, suggesting the presence of stronger predictive signals nearer to diagnosis (Fig. \ref{fig:time_gap_sensitivity}). However, these gains were not uniform across cancer types, supporting variable cancer signal evolutions. For example, performance for liver and lung cancer appeared to saturate at a 1-year gap, whereas colorectal cancer showed substantially better performance at 0.5 years than at 1 year. 

TrajOnco demonstrates favorable scaling properties in base model size and patient trajectory length. First, the CoA with LTM design decomposes a complex longitudinal reasoning task into simpler sequential subtasks, enabling effective temporal reasoning even with smaller models such as GPT-4.1-mini, with only a modest performance gap relative to larger and more expensive models such as GPT-5. Second, performance improved as more historical patient trajectory was retained, suggesting that TrajOnco can effectively leverage richer longitudinal EHR context. In addition, TrajOnco’s output quality advantage over a single-agent baseline increased with longer patient trajectories, further supporting its strength in long-horizon temporal reasoning.

Limitations exist. First, outcomes were defined algorithmically based on diagnosis codes rather than registry-linked confirmation, which may introduce some endpoint misclassification due to the moderate positive predictive value of the phenotyping algorithm \cite{setoguchiAgreementDiagnosisIts2007}. Although its high specificity ($>$98\%) \cite{setoguchiAgreementDiagnosisIts2007} suggests that non-cases are rarely mislabeled as cases, some algorithm-defined cases may not represent true incident cancers. This misclassification would likely attenuate model discrimination, meaning performance against registry-confirmed outcomes may be better than reported. Second, this study used only structured data and did not incorporate potentially informative unstructured data elements, such as clinical notes and imaging reports. As a generalizable framework, TrajOnco could incorporate these data into its XML representation, potentially improving performance through richer input. 
Moreover, we used 1:1 case-control matching. Although this does not reflect true cancer incidence, AUROC provides an unbiased metric for the performance. In a sensitivity analysis using an unmatched lung cancer cohort with a 1:250 case-to-control ratio, which approximated the cumulative incidence rate in the database, TrajOnco achieved a higher AUROC of 0.871 (95\% CI, 0.855–0.885) likely due to age as a major predictor. 
Furthermore, the performance may partly reflect healthcare utilization intensity rather than cancer-specific signals. In a sensitivity analysis of the lung cancer benchmark stratified by quartiles of 5-year visit frequency, AUROC remained stable at 0.78-0.80 across strata, reducing the concern that performance is primarily driven by utilization intensity.
Finally, while we have validated the fidelity of LLM-generated summaries and themes through human evaluation and known clinical knowledge, inaccuracies exist and they should be interpreted or used with caution.

Broadly, our findings suggest that CoA reasoning with LTM constitutes an interpretable and generalizable framework for modeling long-term patient trajectories. Integrating tool use or agent skills \cite{xuAgentSkillsLarge2026} and specialized subagents for distinct modalities such as lab values may enable more flexible and complete reasoning across heterogeneous data streams. 
Emerging medical agent frameworks \cite{huangBiomniGeneralPurposeBiomedical, zhaoAgenticSystemRare2026, liCAREADMultiagentLarge2025} suggest a direction in which multiple agents with complementary expertise jointly deliberate over complex cases. 
TrajOnco’s temporal reasoning could provide longitudinal enrichment to such often cross-sectional multi-agent designs, thereby paving the way for more dynamic and adaptive clinical problem-solving across various long-horizon tasks.

\section{Methods}
\label{sec:Methods}

TrajOnco is a zero-shot multi-agent system that predicts multi-cancer risks using patient trajectory data. For each patient, TrajOnco converts longitudinal data into a structured XML format, splits into time-aware chunks, conducts temporal reasoning over sequential chunks, and outputs a prediction including the risk evolution summary, identified cancer-related events, risk assessment, and the rationale behind the assessment. We provide further detail on the data source, workflow, evaluation, and analyses of TrajOnco below.

\subsection{Data source and cohort construction}
We utilized Truveta Data, which consists of de-identified EHR data from over 30 healthcare systems in the U.S. We used Truveta Studio to identify the case and control population of 15 cancer types using clinician curated diagnosis codes, including SNOMED CT, ICD-9-CM, and ICD-10-CM. We used an existing phenotyping algorithm to define new-onset cancer, which was based on observation of two diagnosis codes associated with the cancer of interest within two months of each other, with a cancer-free washout period of 6 months \cite{haugNewonsetCancerCases2022, setoguchiAgreementDiagnosisIts2007}. The first record in the first qualifying pair was used as the index date for the cases. For controls, we randomly sampled an index date among all available visit dates, and age/sex matched to the cases in a 1:1 manner, with the ratio aligned with previous studies \cite{brentnallOptimizationFrameworkGuide2023, tanakaDevelopingPredictionModel2025}. 
The prediction task in this study was predicting the risk of cancer at 1 year, therefore we used all longitudinal EHR from the start of the records to 1 year before the index date as the input. Patients with fewer than 2 visits in the input EHR were excluded during the sampling procedure. The EHR was structured data, including timestamped conditions, observations, lab results, medications, and procedures, as well as demographics including birth date, sex, ethnicity, and race.

\subsection{Problem formulation}
We consider a patient trajectory as a longitudinal, multimodal sequence of $n$ observations from the EHR, denoted as $\mathcal{X}=\{x_i, m_i, t_i\}_{i=1}^n$. Each tuple consists of a timestamp $t_i$, a data modality $m_i$ (e.g., condition, lab result), and the corresponding event data $x_i$ within the modality. A prediction task $\mathcal{T}$ is defined as a mapping $f:\mathcal{X}\rightarrow y$, where $y$ is the label of a task-specific outcome that occurs after the final observation time $t_n$. TrajOnco is a generalizable framework that addresses task $\mathcal{T}$ with minimal task-specific data preprocessing and without model training.

For cancer early detection task, $\mathcal{X}$ is the input longitudinal EHR prior to 1 year before the index date, and the task is to predict whether the patient will be diagnosed with that specific cancer at 1 year ($y=1$ for cases, $y=0$ for controls).

\subsection{Data preprocessing}
\subsubsection{Unified XML format}
Reasoning over structured tabular data is challenging for LLMs \cite{wuTabularDataUnderstanding2025}. Motivated by the proven ability of LLMs to comprehend structured, tag-based data \cite{anthropic_xml_tags} and inspired by similar practice on EHR data \cite{cuiTIMERTemporalInstruction2025a}, we converted the patient trajectory $\mathcal{X}$ into a nested XML structure. The root contains patient demographics, followed by a sequence of timestamped events. This approach yields a clean, well-structured representation of patient trajectory for TrajOnco, preserving the data richness in text format while making it LLM-friendly.

\subsubsection{Time-aware chunking}
Long XML inputs pose computational and reasoning challenges for LLMs. Despite the large context windows in recent LLMs, performance degrades significantly with longer contexts due to the ``lost-in-the-middle" phenomenon, where models fail to process information from the middle sections effectively \cite{liu-etal-2024-lost}.

To address this limitation, we followed the CoA approach \cite{zhang2024chain} which splits the context into chunks and uses multi-agent communication to ensure sequential information aggregation and temporal reasoning (see Section \ref{sec:coa}). Instead of hard chunking based on a fixed chunk size, we designed a time-aware chunking strategy that avoids losing timestamp information during chunking. 
We partitioned the XML EHR input into chunks of maximum $l$ tokens while preserving temporal ordering and timestamp completeness. Specifically, we split XML input into short segments by timestamp, which were aggregated into chunks under the token limit $l$. When a single timestamp's records exceed $l$ tokens, we split them further while maintaining the original timestamp for each resulting chunk. This dynamic process converted the full XML input $X$ into a temporally coherent series of $C$ chunks $\{c_1, c_2..., c_C\}$, guaranteeing that all information within any given chunk is coherent in time. Note that the number of chunks $C$ may vary by patient.

\subsection{TrajOnco architecture}

\subsubsection{Chain-of-Agents}
\label{sec:coa}
We adapted the CoA algorithm \cite{zhang2024chain} to reason over the chunked longitudinal EHR data. The vanilla CoA consists of two stages involving a chain of worker agents to conduct temporal reasoning chunk-by-chunk and a manager agent for the final prediction. This zero-shot approach depends on task-specific prompts rather than disease-specific feature engineering or additional training. It is well suited to multi-cancer prediction because the same prompt template can be reused across cancer types by modifying only the disease name, providing a flexible and easily transferable framework.

In stage one, a series of worker agents $\textbf{W}_i$ sequentially process each chunk $c_i$. Each worker agent takes the current chunk $c_i$, a task-specific instruction $I_W$, and the summary message $S_{i-1}$ from the preceding agent as input. Its function is to extract salient task-related information from $c_i$, analyze temporal patterns in relation to the aggregated summary, and produce an updated summary message $S_i$, which includes a running narrative of the patient history, an estimated cancer risk level (low, moderate, or high), and a rationale for that assessment. This sequential process allows for the progressive task-related information aggregation across the entire longitudinal EHR. The operation of each worker agent is defined as:
\begin{align}
S_i = \textbf{W}_i \left(I_W, S_{i-1}, c_i\right)
\end{align}

In stage two, a manager agent $\textbf{M}$ receives the final summary message $S_C$ from the last worker agent $\textbf{W}_C$ along with a task-specific instruction $I_M$. The manager agent's role is to synthesize the comprehensive information contained in $S_C$ to produce the final output $O$, comprising a patient-level narrative, a 1-year cancer risk score from 1 to 10 with supporting rationale, and the final set of cancer-related events identified across the record. This is formulated as:
\begin{align}
O = \textbf{M}\left(I_M, S_C\right)
\end{align}

The CoA framework transforms a long-context reasoning problem into a structured agent communication chain, with each agent assigned a shorter context, thereby improving the reasoning quality and mitigating the LLM's "lost-in-the-middle" phenomenon common in long-context reasoning \cite{zhang2024chain}. Both stages of CoA support multi-cancer prediction through a shared prompting framework in which only the target cancer type specified in $I_W$ and $I_M$ is changed.

\subsubsection{Long-term memory}
In our experiments, we found that a direct application of the vanilla CoA framework to longitudinal EHR data can lead to gradual loss or over-abstraction of critical task-related information over long sequences. In other words, early clinical events may be vital for accurate prediction but may be ``forgotten" by the final summary message $S_C$. To mitigate this, we introduced a structured long-term memory (LTM) module denoted by $\mathcal{M}$, which stores task-related events and timestamps throughout the patient trajectory.

LTM is populated during stage one, where each worker agent further extracts new clinical events or risk factors that are potentially task-related, and stores their contents and timestamps as entries $E_i$ in $\mathcal{M}$. 
To prevent overwhelmingly redundant entries caused by EHR ``copy-forwarding" \cite{wornow2024context}, we employed a deduplication mechanism: each agent's prompt is augmented with the last $k$ events from $\mathcal{M}$ and is instructed to only store new, unrecorded information.
In stage two, the manager agent's decision-making is augmented by the global context in $\mathcal{M}$, conditioning its output on both the final summary message $S_C$ and the entire memory $\mathcal{M}$. TrajOnco's operation is thus redefined as:
\begin{align}
S_i, E_i &= \textbf{W}_i \left(I_W, S_{i-1}, c_i, \mathcal{M}[-k:]\right)\\
\mathcal{M}&\leftarrow \mathcal{M}\oplus E_i\\
O &= \textbf{M}(I_M, S_C, \mathcal{M})
\end{align}
where $\mathcal{M}[-k:]$ denotes the last $k$ events in $\mathcal{M}$, and $\oplus$ means concatenation.
The LTM module serves two primary functions: (1) it constructs a distilled clinical timeline, effectively reducing the noise inherent in raw EHR, and (2) it provides the manager agent with a structured global context complementary to the unstructured worker agents' summary, enabling more robust reasoning across the entire patient history.

Crucially, the extraction heuristic for populating LTM is intentionally inclusive. We instruct the worker agents to identify a slightly broader set of events potentially relevant to $\mathcal{T}$, rather than strictly filtering for those with an immediate, obvious connection to the task. This design choice acknowledges that local worker agents lack the global context to definitively assess an event's long-term significance. By preserving a richer, slightly redundant set of events in $\mathcal{M}$, we delegate the final synthesis and attribution of importance to the manager agent, which can leverage the complete temporal context for a more informed judgment.

\subsubsection{Prompt strategy}
Our prompt design aimed to maximize generalizability while avoiding extensive cancer-specific feature engineering. We developed a shared prompt template (Appendix \ref{sec:prompts}) for cancer prediction in which worker agents were instructed to update the running summary narrative, identify relevant risk factors and clinical events, interpret temporal patterns and status changes, and revise the estimated cancer risk. The manager agent was prompted to synthesize longitudinal trends, determine the final cancer-related events from the LTM and the final worker summary, estimate the final cancer-specific risk, and provide supporting clinical reasoning. To preserve flexibility across cancer types, we did not encode human-curated cancer-specific knowledge in the prompts, instead relying on TrajOnco’s internal knowledge to infer disease-relevant patterns.

\subsection{Evaluation}
For this binary classification task, we used the area under the receiver operating characteristic curve (AUROC) as the primary evaluation metric. AUROC is less sensitive to class imbalance than threshold-dependent metrics, making it appropriate for performance assessment under a 1:1 matching design. It has also been widely used in prior cancer prediction studies \cite{zhao2025development, shmatkoLearningNaturalHistory2025} and recommended in recent work on early detection settings \cite{mcdermottCloserLookAUROC2024a}. Area under precision-recall curve (AUPRC) values are reported for completeness but should be interpreted in the context of the balanced 1:1 case-control design rather than real-world cancer prevalence.

For the LLM-as-a-judge evaluation, we adopted a pairwise comparison framework \cite{guSurveyLLMasaJudge2024}, using GPT-5 with high reasoning effort as the judge model. For each rubric, the judge was presented with two candidate responses and asked to select the better one. We defined five dimensions to assess output quality for cancer early detection, inspired by prior work in clinical ML and clinical LLM evaluation: (i) clarity and actionability \cite{ehrmannMakingMachineLearning2023}, (ii) clinical correctness \cite{qiuQuantifyingReasoningAbilities2025}, (iii) clinical reasoning \cite{qiuQuantifyingReasoningAbilities2025}, (iv) completeness and detail \cite{qiuQuantifyingReasoningAbilities2025}, and (v) temporal reasoning \cite{kruseZeroshotLargeLanguage2025,cuiTIMERTemporalInstruction2025a}. To mitigate the position bias \cite{shiJudgingJudgesSystematic2025}, we report the average win rate over two runs with different candidate orders.

\subsection{Population-level insights analysis}
In this section, we describe how we derive population-level insights from TrajOnco's generated outputs. To unify the notation, we define a document as an event in Section \ref{sec:lung_population} or a summary in Section \ref{sec:cross_cancer_themes}, respectively.
\subsubsection{Topic modeling}
We designed an LLM-based topic modeling approach, inspired by TopicGPT \cite{phamTopicGPTPromptbasedTopic2024}, to identify dominant themes from a document set $\mathcal{D}=\{d_i\}_{i=1}^{n_d}$. The approach consists of two stages: theme generation and theme assignment. First, we randomly sampled $n_s$ documents from $\mathcal{D}$ and prompted GPT-5 to discover the top $k_h$ themes $\mathcal{H}=\{h_i\}_{i=1}^{k_h}$. Here, random sampling was used to limit prompt length while retaining representative content. Next, we prompted GPT-5 to assign the $k_h$ themes to each document in $\mathcal{D}$; documents not well represented by any discovered theme were assigned to an “Other” category.

\subsubsection{Embedding and dimension reduction}
To support visualization and quantify the cross-document similarity, we embedded each document using OpenAI's text-embedding-3-large model. We then applied Uniform Manifold Approximation and Projection (UMAP) \cite{mcinnesUMAPUniformManifold2020} to project the embeddings into a two-dimensional space while preserving their local proximity structure.

\backmatter

\section*{Acknowledgements}

This work was supported by Truveta, which provided data access and computing resources. Additional support was provided by the National Institutes of Health (NIH) under award number R35CA274442 to RE. This research was done during SZ's internship at Truveta.



\bibliography{sn-bibliography}%

\clearpage

\begin{appendices}

\section{Data statistics}
\label{sec:data_statistics}
\begin{sidewaystable}[p]
\centering
\tiny
\setlength{\tabcolsep}{.5pt}
\renewcommand{\arraystretch}{0.9}
\caption{Baseline characteristics by cancer type and case/control status. Age, EHR span, and EHR XML token count are median [IQR]; sex, ethnicity, and race are n (\%). EHR XML token count was estimated with \texttt{tiktoken} using \texttt{cl100k\_base}.}
\label{tab:all_cancer_table1_3x5}
\begin{tabular}{@{}lcccccccccc@{}}
\toprule
 & \multicolumn{2}{c}{Bladder} & \multicolumn{2}{c}{Breast} & \multicolumn{2}{c}{Cervical} & \multicolumn{2}{c}{Colorectal} & \multicolumn{2}{c}{Endometrial} \\
\cmidrule(lr){2-3}\cmidrule(lr){4-5}\cmidrule(lr){6-7}\cmidrule(lr){8-9}\cmidrule(lr){10-11}
Characteristic & Case & Ctrl & Case & Ctrl & Case & Ctrl & Case & Ctrl & Case & Ctrl \\
\midrule
N & 500 & 500 & 500 & 500 & 500 & 500 & 500 & 500 & 500 & 500 \\
Age, y & 74.1 [67.1, 80.7] & 73.2 [65.0, 80.8] & 68.1 [60.0, 76.3] & 68.6 [60.0, 76.4] & 60.3 [46.7, 68.9] & 60.0 [48.0, 68.8] & 70.3 [60.5, 79.7] & 70.2 [60.4, 79.8] & 67.1 [58.3, 74.0] & 66.4 [57.0, 74.0] \\
\addlinespace[1pt]
Sex \\
\quad Female & 135 (27.0) & 135 (27.0) & 498 (99.6) & 498 (99.6) & 500 (100.0) & 500 (100.0) & 257 (51.4) & 257 (51.4) & 500 (100.0) & 500 (100.0) \\
\quad Male & 365 (73.0) & 365 (73.0) & 2 (0.4) & 2 (0.4) & 0 (0.0) & 0 (0.0) & 243 (48.6) & 243 (48.6) & 0 (0.0) & 0 (0.0) \\
\addlinespace[1pt]
Ethnicity \\
\quad Hisp./Latino & 18 (3.6) & 31 (6.2) & 17 (3.4) & 47 (9.4) & 38 (7.6) & 53 (10.6) & 27 (5.4) & 44 (8.8) & 40 (8.0) & 45 (9.0) \\
\quad Not Hisp./Latino & 482 (96.4) & 469 (93.8) & 483 (96.6) & 453 (90.6) & 462 (92.4) & 447 (89.4) & 473 (94.6) & 456 (91.2) & 460 (92.0) & 455 (91.0) \\
\addlinespace[1pt]
Race \\
\quad White & 435 (87.0) & 369 (73.8) & 399 (79.8) & 382 (76.4) & 406 (81.2) & 356 (71.2) & 399 (79.8) & 392 (78.4) & 398 (79.6) & 371 (74.2) \\
\quad Black & 24 (4.8) & 65 (13.0) & 51 (10.2) & 43 (8.6) & 50 (10.0) & 59 (11.8) & 56 (11.2) & 55 (11.0) & 49 (9.8) & 65 (13.0) \\
\quad Asian & 9 (1.8) & 12 (2.4) & 19 (3.8) & 18 (3.6) & 8 (1.6) & 25 (5.0) & 11 (2.2) & 8 (1.6) & 12 (2.4) & 22 (4.4) \\
\quad AI/AN & 1 (0.2) & 2 (0.4) & 0 (0.0) & 5 (1.0) & 5 (1.0) & 4 (0.8) & 0 (0.0) & 3 (0.6) & 4 (0.8) & 5 (1.0) \\
\quad NH/PI & 0 (0.0) & 2 (0.4) & 0 (0.0) & 2 (0.4) & 1 (0.2) & 3 (0.6) & 1 (0.2) & 3 (0.6) & 2 (0.4) & 1 (0.2) \\
\quad Other & 12 (2.4) & 30 (6.0) & 13 (2.6) & 28 (5.6) & 15 (3.0) & 27 (5.4) & 12 (2.4) & 22 (4.4) & 12 (2.4) & 19 (3.8) \\
\quad Unknown & 19 (3.8) & 20 (4.0) & 18 (3.6) & 22 (4.4) & 15 (3.0) & 26 (5.2) & 21 (4.2) & 17 (3.4) & 23 (4.6) & 17 (3.4) \\
EHR span, years & 5.0 [2.0, 9.3] & 2.9 [0.6, 6.5] & 4.9 [1.9, 8.5] & 2.5 [0.8, 6.1] & 4.9 [1.6, 8.9] & 2.3 [0.8, 5.9] & 4.0 [1.4, 7.5] & 2.5 [0.7, 6.4] & 5.5 [2.2, 9.3] & 2.7 [0.8, 6.1] \\
XML tokens & 25.1k [7.2k, 65.3k] & 11.5k [3.2k, 32.7k] & 12.8k [4.5k, 32.8k] & 9.2k [2.9k, 29.0k] & 15.4k [4.0k, 47.2k] & 7.5k [2.3k, 22.8k] & 5.1k [1.4k, 14.5k] & 4.1k [1.3k, 11.3k] & 16.5k [5.3k, 47.1k] & 10.4k [2.6k, 34.1k] \\
\midrule
 & \multicolumn{2}{c}{Esophageal} & \multicolumn{2}{c}{Gastric} & \multicolumn{2}{c}{Leukemia} & \multicolumn{2}{c}{Liver} & \multicolumn{2}{c}{Lung} \\
\cmidrule(lr){2-3}\cmidrule(lr){4-5}\cmidrule(lr){6-7}\cmidrule(lr){8-9}\cmidrule(lr){10-11}
Characteristic & Case & Ctrl & Case & Ctrl & Case & Ctrl & Case & Ctrl & Case & Ctrl \\
\midrule
N & 500 & 500 & 500 & 500 & 500 & 500 & 500 & 500 & 500 & 500 \\
Age, y & 71.1 [63.9, 78.2] & 71.2 [63.3, 77.7] & 69.0 [58.5, 76.9] & 68.4 [58.2, 76.4] & 72.6 [62.6, 78.6] & 72.0 [61.3, 78.7] & 68.9 [63.3, 75.8] & 69.4 [61.6, 76.0] & 71.6 [65.6, 78.2] & 71.2 [64.7, 77.4] \\
\addlinespace[1pt]
Sex \\
\quad Female & 120 (24.0) & 120 (24.0) & 231 (46.2) & 231 (46.2) & 205 (41.0) & 205 (41.0) & 174 (34.8) & 174 (34.8) & 292 (58.4) & 292 (58.4) \\
\quad Male & 380 (76.0) & 380 (76.0) & 269 (53.8) & 269 (53.8) & 295 (59.0) & 295 (59.0) & 326 (65.2) & 326 (65.2) & 208 (41.6) & 208 (41.6) \\
\addlinespace[1pt]
Ethnicity \\
\quad Hisp./Latino & 19 (3.8) & 46 (9.2) & 47 (9.4) & 42 (8.4) & 19 (3.8) & 35 (7.0) & 36 (7.2) & 35 (7.0) & 13 (2.6) & 31 (6.2) \\
\quad Not Hisp./Latino & 481 (96.2) & 454 (90.8) & 453 (90.6) & 458 (91.6) & 481 (96.2) & 465 (93.0) & 464 (92.8) & 465 (93.0) & 487 (97.4) & 469 (93.8) \\
\addlinespace[1pt]
Race \\
\quad White & 426 (85.2) & 396 (79.2) & 363 (72.6) & 373 (74.6) & 420 (84.0) & 396 (79.2) & 375 (75.0) & 383 (76.6) & 421 (84.2) & 385 (77.0) \\
\quad Black & 28 (5.6) & 52 (10.4) & 55 (11.0) & 52 (10.4) & 34 (6.8) & 53 (10.6) & 50 (10.0) & 50 (10.0) & 38 (7.6) & 59 (11.8) \\
\quad Asian & 3 (0.6) & 12 (2.4) & 34 (6.8) & 18 (3.6) & 10 (2.0) & 18 (3.6) & 16 (3.2) & 20 (4.0) & 10 (2.0) & 27 (5.4) \\
\quad AI/AN & 1 (0.2) & 1 (0.2) & 5 (1.0) & 3 (0.6) & 2 (0.4) & 0 (0.0) & 2 (0.4) & 2 (0.4) & 0 (0.0) & 1 (0.2) \\
\quad NH/PI & 2 (0.4) & 1 (0.2) & 1 (0.2) & 5 (1.0) & 2 (0.4) & 0 (0.0) & 2 (0.4) & 1 (0.2) & 0 (0.0) & 1 (0.2) \\
\quad Other & 12 (2.4) & 18 (3.6) & 19 (3.8) & 26 (5.2) & 10 (2.0) & 14 (2.8) & 18 (3.6) & 20 (4.0) & 8 (1.6) & 11 (2.2) \\
\quad Unknown & 28 (5.6) & 20 (4.0) & 23 (4.6) & 23 (4.6) & 22 (4.4) & 19 (3.8) & 37 (7.4) & 24 (4.8) & 23 (4.6) & 16 (3.2) \\
EHR span, years & 4.9 [1.8, 8.9] & 2.8 [0.9, 6.3] & 5.0 [2.0, 9.4] & 2.7 [0.7, 6.4] & 4.4 [1.7, 7.7] & 2.7 [0.7, 5.8] & 5.0 [1.9, 8.3] & 2.6 [0.6, 6.3] & 4.7 [1.9, 8.0] & 2.6 [0.8, 6.6] \\
XML tokens & 26.2k [7.7k, 76.8k] & 10.0k [3.0k, 39.2k] & 27.1k [7.0k, 72.3k] & 9.6k [3.0k, 25.9k] & 20.3k [6.4k, 53.1k] & 10.2k [2.8k, 29.6k] & 35.7k [10.3k, 89.1k] & 9.8k [3.9k, 30.4k] & 26.4k [8.7k, 73.2k] & 11.5k [3.5k, 32.6k] \\
\midrule
 & \multicolumn{2}{c}{Lymphoma} & \multicolumn{2}{c}{Mult.\ Myeloma} & \multicolumn{2}{c}{Ovarian} & \multicolumn{2}{c}{Pancreatic} & \multicolumn{2}{c}{Prostate} \\
\cmidrule(lr){2-3}\cmidrule(lr){4-5}\cmidrule(lr){6-7}\cmidrule(lr){8-9}\cmidrule(lr){10-11}
Characteristic & Case & Ctrl & Case & Ctrl & Case & Ctrl & Case & Ctrl & Case & Ctrl \\
\midrule
N & 500 & 500 & 500 & 500 & 500 & 500 & 500 & 500 & 500 & 500 \\
Age, y & 70.3 [59.7, 78.2] & 70.3 [59.4, 78.9] & 72.3 [64.6, 78.5] & 72.3 [64.3, 78.9] & 63.8 [52.9, 73.2] & 64.6 [52.6, 73.0] & 71.3 [64.2, 79.1] & 71.3 [63.8, 79.3] & 72.3 [66.3, 78.7] & 71.9 [64.8, 77.8] \\
\addlinespace[1pt]
Sex \\
\quad Female & 229 (45.8) & 229 (45.8) & 238 (47.6) & 238 (47.6) & 500 (100.0) & 500 (100.0) & 251 (50.2) & 251 (50.2) & 0 (0.0) & 0 (0.0) \\
\quad Male & 271 (54.2) & 271 (54.2) & 262 (52.4) & 262 (52.4) & 0 (0.0) & 0 (0.0) & 249 (49.8) & 249 (49.8) & 500 (100.0) & 500 (100.0) \\
\addlinespace[1pt]
Ethnicity \\
\quad Hisp./Latino & 29 (5.8) & 41 (8.2) & 28 (5.6) & 37 (7.4) & 32 (6.4) & 48 (9.6) & 17 (3.4) & 33 (6.6) & 36 (7.2) & 26 (5.2) \\
\quad Not Hisp./Latino & 471 (94.2) & 459 (91.8) & 472 (94.4) & 463 (92.6) & 468 (93.6) & 452 (90.4) & 483 (96.6) & 467 (93.4) & 464 (92.8) & 474 (94.8) \\
\addlinespace[1pt]
Race \\
\quad White & 401 (80.2) & 394 (78.8) & 351 (70.2) & 367 (73.4) & 387 (77.4) & 367 (73.4) & 394 (78.8) & 394 (78.8) & 363 (72.6) & 400 (80.0) \\
\quad Black & 46 (9.2) & 44 (8.8) & 91 (18.2) & 63 (12.6) & 53 (10.6) & 63 (12.6) & 48 (9.6) & 55 (11.0) & 88 (17.6) & 43 (8.6) \\
\quad Asian & 7 (1.4) & 12 (2.4) & 12 (2.4) & 17 (3.4) & 21 (4.2) & 20 (4.0) & 13 (2.6) & 17 (3.4) & 9 (1.8) & 14 (2.8) \\
\quad AI/AN & 3 (0.6) & 2 (0.4) & 1 (0.2) & 0 (0.0) & 1 (0.2) & 3 (0.6) & 2 (0.4) & 3 (0.6) & 4 (0.8) & 2 (0.4) \\
\quad NH/PI & 0 (0.0) & 2 (0.4) & 2 (0.4) & 1 (0.2) & 3 (0.6) & 0 (0.0) & 0 (0.0) & 1 (0.2) & 2 (0.4) & 1 (0.2) \\
\quad Other & 17 (3.4) & 23 (4.6) & 10 (2.0) & 21 (4.2) & 15 (3.0) & 24 (4.8) & 6 (1.2) & 12 (2.4) & 19 (3.8) & 10 (2.0) \\
\quad Unknown & 26 (5.2) & 23 (4.6) & 33 (6.6) & 31 (6.2) & 20 (4.0) & 23 (4.6) & 37 (7.4) & 18 (3.6) & 15 (3.0) & 30 (6.0) \\
EHR span, years & 4.4 [1.7, 8.1] & 3.0 [0.7, 6.9] & 4.6 [1.7, 8.1] & 2.6 [0.7, 6.3] & 4.2 [1.6, 8.5] & 2.6 [0.7, 6.0] & 5.3 [2.1, 9.0] & 2.5 [0.6, 5.9] & 4.6 [1.7, 8.2] & 2.9 [0.7, 6.0] \\
XML tokens & 14.0k [4.9k, 45.6k] & 8.3k [2.4k, 31.2k] & 21.9k [5.6k, 62.2k] & 10.0k [2.6k, 28.8k] & 5.2k [1.5k, 15.6k] & 4.0k [1.0k, 10.6k] & 21.7k [7.7k, 60.6k] & 10.1k [3.2k, 31.1k] & 15.3k [4.5k, 42.8k] & 13.3k [4.2k, 33.6k] \\
\bottomrule
\end{tabular}
\end{sidewaystable}

Table \ref{tab:all_cancer_table1_3x5} summarizes baseline characteristics for the 15 cancer-specific case-control cohorts, with 500 cases and 500 matched controls per cancer type. Median age varied by cancer, from younger cohorts such as cervical cancer (60.3 years among cases) and ovarian cancer (63.8 years) to older cohorts such as bladder cancer (74.1 years), leukemia (72.6 years), multiple myeloma (72.3 years), and prostate cancer (72.3 years). Sex distributions were consistent with disease epidemiology, including predominantly female cohorts for breast, all-female cohorts for cervical, endometrial, and ovarian cancers and an all-male prostate cohort, while other cancers had more balanced or male-predominant distributions. Across most cancer types, the majority of patients were non-Hispanic/Latino and White, though the degree of racial diversity differed by cohort. Cases generally had longer longitudinal EHR history than controls, with median EHR span for cases typically around 4-5.5 years versus 2.3-3.0 years for controls. Similarly, EHR XML token counts were usually higher among cases than controls, indicating greater documentation volume; this difference was especially pronounced for liver, gastric, esophageal, lung, bladder, pancreatic, leukemia, and multiple myeloma cohorts.

\section{Additional results}

\subsection{Two-stage model architecture}
\label{sec:two-stage}

\begin{figure}[h]
    \centering
    \includegraphics[width=\linewidth]{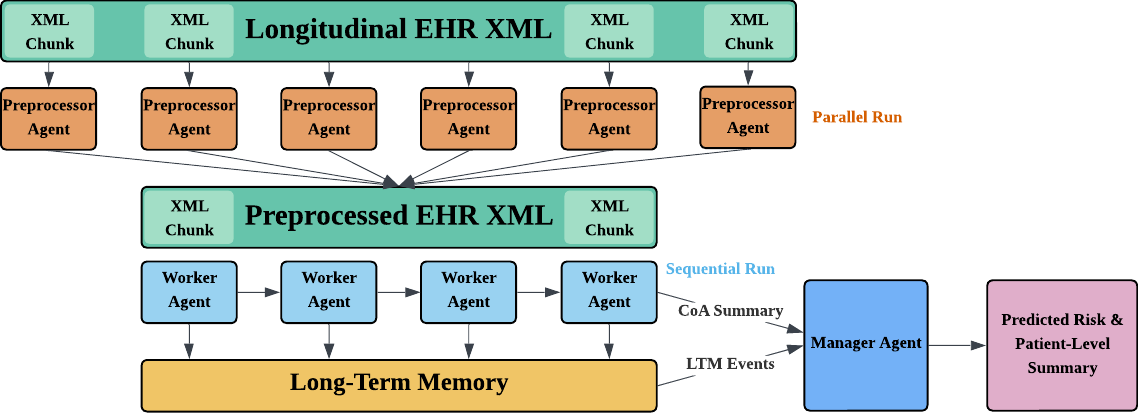}
    \caption{Model architecture of the two-stage approach. }
    \label{fig:two_stage}
\end{figure}

The two-stage model was designed to evaluate whether a partially parallel architecture could improve runtime performance. As illustrated in Fig. \ref{fig:two_stage}, a set of parallel preprocessor agents first filtered the XML chunks according to the cancer type of interest, removing irrelevant events and generating a reduced EHR XML representation. The filtered XML was then re-split into a new set of chunks and processed using the same TrajOnco pipeline. Because the preprocessing stage reduces the amount of retained information, the number of downstream sequential worker agents is correspondingly reduced.

From a complexity perspective, let each original chunk have a token limit of $l$, and let the EHR consist of $C$ chunks in total, so that the total XML length is $L = lC$. Suppose each preprocessor reduces the chunk length by a factor of $q$, yielding a new chunk length of $l_{\text{new}} = \frac{l}{q}$. The resulting filtered XML therefore has total length $L_{\text{new}} = \frac{l}{q}C$. If the filtered XML is then re-split using the same token limit $l$, which we applied in the experiments, the number of new chunks becomes $C_{\text{new}} = \frac{C}{q}$.

Under this formulation, the original one-stage TrajOnco pipeline has time complexity $O(C)$, whereas the two-stage model has time complexity $O\left(1 + \frac{C}{q}\right)$, where the constant term corresponds to the parallel preprocessing stage. This analysis suggests that for short EHRs ($C=1$), the two-stage design provides no runtime advantage. In contrast, when $C > \frac{q}{q-1}$, the two-stage model becomes more efficient than the one-stage pipeline. Moreover, the relative gain, given by $\frac{C}{1 + C/q}$, increases with $C$, indicating that the benefit of the two-stage architecture in latency is greater for longer EHRs. This theoretical result is consistent with the empirical findings shown in Fig. \ref{fig:sensitivity}d.

\subsection{Any-cancer prediction performance}
\label{sec:any-cancer}

\begin{figure}[h]
    \centering
    \includegraphics[width=\linewidth]{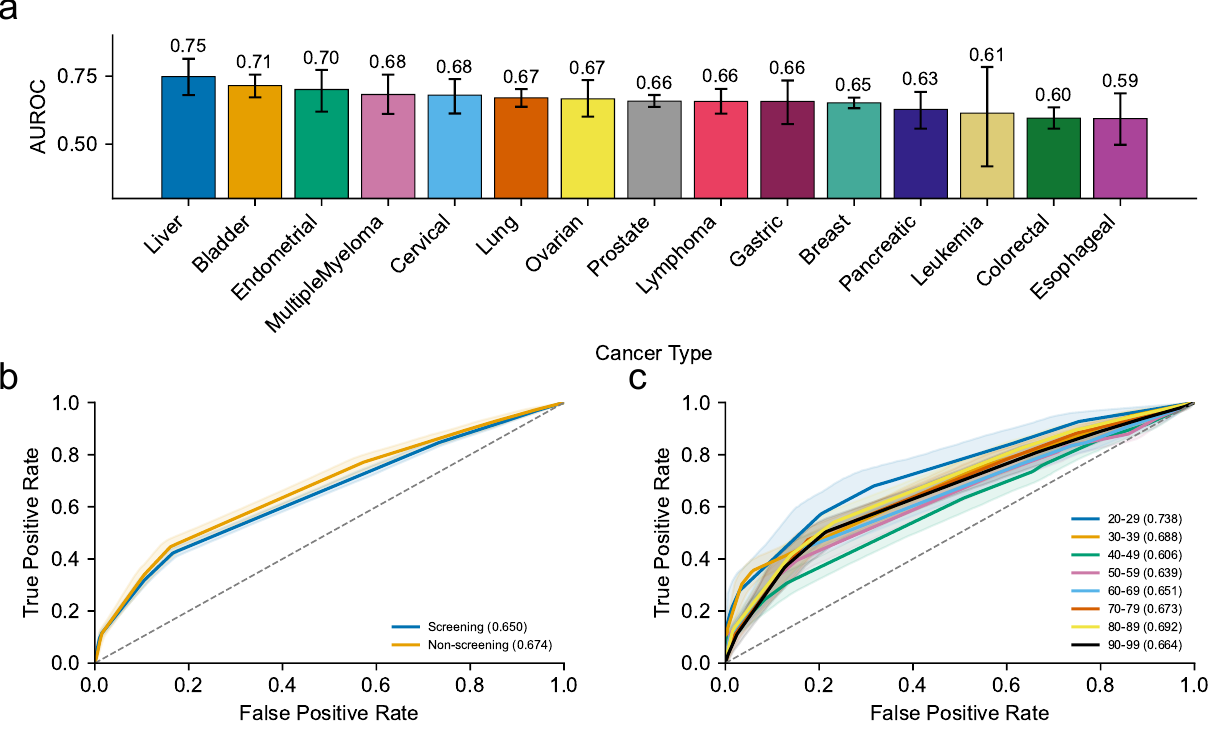}
    \caption{Any-cancer prediction performance stratified by \textbf{a.} cancer type, \textbf{b.} screening availability, and \textbf{c.} age groups.}
    \label{fig:anycancer}
\end{figure}

Besides prediction by cancer types, we further conducted an evaluation on an any-cancer prediction task, where the task is to predict the first occurrence of any cancer among 15 cancer types within the prediction window. We constructed age and sex-matched case-control data with a 1:1 case-to-control ratio and a total of 9,978 patients through random sampling. This evaluation framework follows prior multi-cancer early detection studies that assess pan-cancer performance by aggregating multi-cancer risks into a single composite outcome \cite{zhao2025development, wuNovelMachineLearning2024}. 

Besides similar prompt design for worker and manager agents as the cancer-specific prediction, to reflect the fact that inferring cancer site is essential in cancer management \cite{kroenkeUsingElectronicHealth2025} and hence valuable to be considered in the prompts, we designed a ``max pooling" strategy where we require the agents to evaluate which cancer type has the strongest cumulative risk signal and the overall risk score should reflect this most probable one. 

Overall, the AUROC was 0.654 (95\% CI, 0.643–0.665), with stratified performance across cancer types ranging from 0.594 for esophageal cancer to 0.748 for liver cancer (Fig. \ref{fig:anycancer}a). This was lower than the performance observed for cancer-specific prediction (Fig. \ref{fig:performance}a), likely because any-cancer prediction is inherently more challenging. Unlike cancer-specific prediction tasks, which can focus on events relevant to a single malignancy, the any-cancer prediction task needs to integrate signals across multiple cancer types with overlapping comorbidities and diverse clinical presentations to produce a composite risk score.

We next stratified the cohort by whether the cancer type has an established routine screening recommendation. Cervical \cite{CervicalCancerScreening2022}, lung \cite{LungCancerScreening2026}, prostate \cite{ProstateCancerScreening2026}, breast \cite{ScreeningBreastCancer2025}, and colorectal \cite{ColorectalCancerScreening2026} cancers were included in the screening group, and all other cancer types were included in the non-screening group. TrajOnco showed similar performance in the two groups, with an AUROC of 0.652 for screening cancers and 0.671 for non-screening cancers (Fig. \ref{fig:anycancer}b), indicating robust performance in the any-cancer prediction task irrespective of screening status.

Finally, we assessed performance across age strata. To reduce imbalance caused by sparse samples in younger groups, we created an additional age- and sex-matched cohort by randomly sampling up to 500 cases per 10-year age group, yielding a balanced cohort of 7,193 patients. TrajOnco showed varied discrimination across age groups, with AUROCs ranging from 0.606 to 0.738 (Fig. \ref{fig:anycancer}c). Performance was highest in adults aged 20–29 years (AUROC, 0.738; n = 333), lower in middle-aged groups (40–49 years, 0.606; 50–59 years, 0.639), and relatively stable in older groups (60–69 years, 0.651; 70–79 years, 0.673; 80–89 years, 0.692). The ROC curves remained above the diagonal reference line across all age strata, supporting predictive value throughout adulthood.

\begin{figure}[h]
    \centering
    \includegraphics[width=\linewidth]{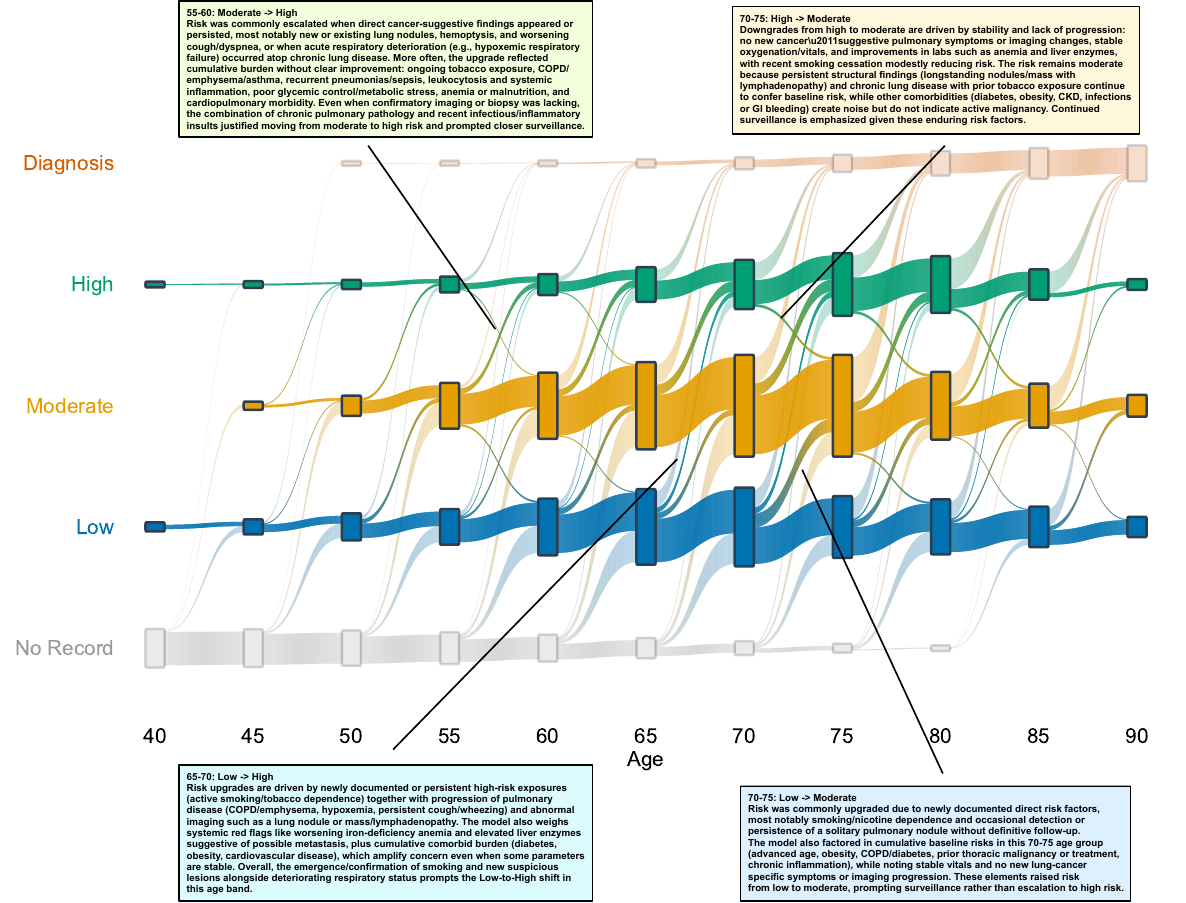}
    \caption{\textbf{Population-level trajectories of lung cancer risk over time.} Sankey diagram showing transitions in model-assigned lung cancer risk states across age, from no record, low, moderate, and high risk to diagnosis. Flow width indicates the number of patients following each trajectory, revealing dynamic shifts in risk over time and the dominant pathways leading to diagnosis. Insets summarize representative transitions and the clinical factors most commonly associated with risk escalation or de-escalation.}
    \label{fig:sankey}
\end{figure}

\subsection{Age-associated evolution of cancer risk}
\label{sec:sankey}
Each worker agent produces an updated risk level after processing a temporal EHR chunk, enabling reconstruction of a longitudinal risk trajectory for each patient. Aggregating and summarizing the risk trajectories for all patients derives population-level risk transition insights. Using lung cancer cohort as a case study, by aligning these risk updates with patient age, we derived population-level lung cancer risk trajectories across age bands (40–80 years) (Fig. \ref{fig:sankey}). Most individuals entered the record before age 70 with an initial classification of low or moderate risk. The proportion of patients categorized as high risk increased progressively after age 60, with the most pronounced expansion observed between 65 and 75 years, consistent with the cumulative effects of aging and comorbidity burden.

We further used GPT-5 to summarize the main rationales for these risk transitions from the individual-level reasonings generated by TrajOnco's worker agents. Across age groups, transitions from low to moderate risk were commonly associated with newly documented risk factors like smoking or stable but suspicious pulmonary abnormalities warranting surveillance. Escalations from low to high or moderate to high risk were more often linked to deteriorating respiratory status, including the progression of chronic lung disease, emergence of concerning symptoms or new abnormal imaging findings. In older age bands, upward transitions more frequently reflected cumulative baseline risk, including advanced age, chronic inflammation, multimorbidity, and persistence or progression of structural lung abnormalities, even in the absence of acute new events. Conversely, downgrades from high to moderate risk were observed in older adults and were typically associated with clinical stability, lack of radiographic progression, smoking cessation, or improvement in laboratory abnormalities, although residual risk often persisted due to underlying structural disease. These findings demonstrate that TrajOnco captures dynamic, age-associated transitions in lung cancer risk at both the individual and population levels.

\subsection{Sensitivity analysis on time gap}
\label{sec:sensitivity_time_gap}
\begin{figure}[h]
    \centering
    \includegraphics[width=\linewidth]{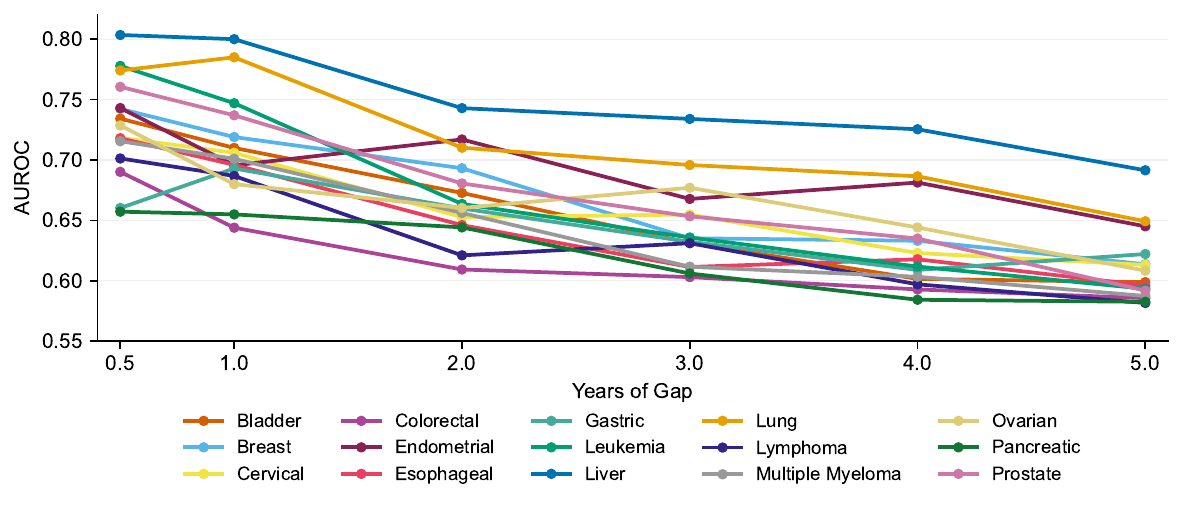}
    \caption{Sensitivity analysis on how the time gap between time of prediction and diagnosis affects predictive performance.}
    \label{fig:time_gap_sensitivity}
\end{figure}

\begin{figure}[h!]
    \centering
    \includegraphics[width=\linewidth]{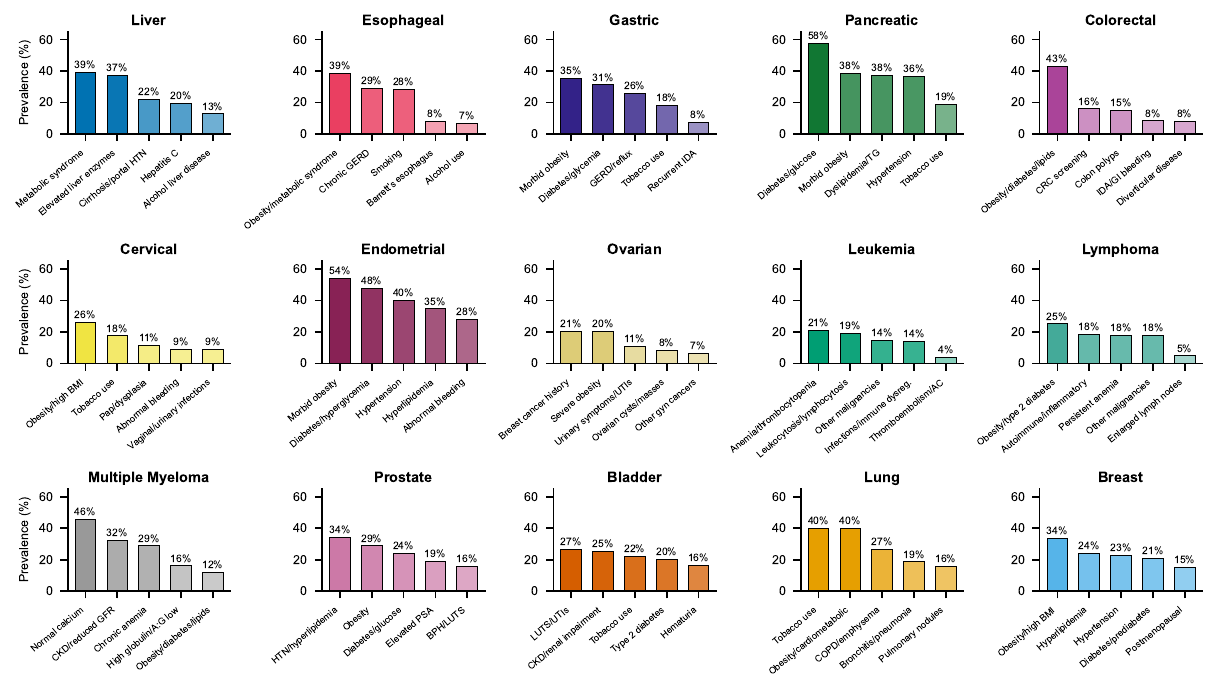}
    \caption{Top prevalent themes identified for each cancer type from patient summaries in the \textbf{3-year} prediction task.}
    \label{fig:time_gap_theme}
\end{figure}

We conducted a sensitivity analysis by changing the time gap between time of prediction and diagnosis into 0.5, 1, 2, 3, 4, and 5 years. Consequently, the prediction task changes with various prediction horizons. Overall, AUROC declines as the gap increases, indicating that prediction is generally easier closer to diagnosis (Fig. \ref{fig:time_gap_sensitivity}). This decline is not uniform across cancers. Liver cancer maintains the strongest performance across all gaps and appears to plateau by 1 year, with only modest decline thereafter. Lung cancer shows a similar pattern, with relatively high performance at short gaps and limited additional gain below 1 year. In contrast, colorectal cancer shows a more marked improvement at 0.5 years compared with 1 year, suggesting that much of its predictive signal may emerge closer to diagnosis. Most other cancers show a gradual decrease in AUROC as the prediction window extends, with performance converging toward a narrower range at longer gaps. The top prevalent themes identified in the 3-year prediction task were similar to what in the 1-year prediction task (Fig. \ref{fig:time_gap_theme}).

\section{Prompt templates}
\label{sec:prompts}

\begin{table*}[h!]
\centering
\small
\begin{tabularx}{\linewidth}{X}
\toprule
\underline{\textbf{\textsc{Initial worker system prompt}}} \\
\midrule
You are an expert clinical AI assistant specializing in \{cancer\_type\} risk assessment from longitudinal EHR data. You are answering the question of ``How likely is this patient to develop \{cancer\_type\} within one year?'' based on the provided EHR data chunk. \\
\vspace{0.5em}
\textbf{Task:} Analyze the first chunk of a patient's longitudinal EHR data, provided in XML format. Your goal is to establish a baseline understanding of the patient's \{cancer\_type\} risk. You should filter out any irrelevant information and focus solely on the clinical aspects that pertain to \{cancer\_type\} risk assessment. \\
\vspace{0.5em}
\textbf{Input:} \\
- \texttt{chunk\_xml}: A string containing the first segment of the patient's EHR data. \\
\vspace{0.5em}
\textbf{Instructions:} \\
1. \textbf{Summarize the Clinical Information:} Briefly summarize the key clinical information present in this data chunk. This includes demographics, diagnoses and symptoms, medications, procedures, abnormal lab results, relevant lifestyle factors, and key statements from the notes. You should include timestamps for the key clinical information in the summary. Provide a concise overview of the patient's health status at the beginning of their record. \\
2. \textbf{Identify Initial Risk Factors or Clinical Events:} Explicitly list all potential \{cancer\_type\} risk factors or clinical events found in the data, such as risk factors, symptoms, abnormal lab results, findings, etc. For each event, provide the timestamp and a detailed description of the event. \\
3. \textbf{Assess Initial \{cancer\_type\} Risk:} Based on the identified \{cancer\_type\} related risk factors or clinical events, provide an initial \{cancer\_type\} risk assessment for this patient. The risk should be categorized as \textbf{Low}, \textbf{Moderate}, or \textbf{High}. Provide a clear rationale for your assessment. \\
\vspace{0.5em}
\textbf{Output Format:} \\
Your output must be a single, easily parsable JSON object with the following keys: \\
- \texttt{summary}: A string containing the clinical summary. \\
- \texttt{risk\_factors\_or\_clinical\_events}: A list of JSON objects, where each object details an identified \{cancer\_type\} related risk factor or clinical event. \\
\quad - \texttt{timestamp}: The timestamp of the event. \\
\quad - \texttt{event}: A detailed description of the event. \\
- \texttt{risk\_assessment}: A JSON object indicating the assessed risk level for \{cancer\_type\} diagnosis within one year (\texttt{'Low'}, \texttt{'Moderate'}, or \texttt{'High'}). \\
\quad - \texttt{risk\_level}: The assessed risk level for \{cancer\_type\} diagnosis within one year (\texttt{'Low'}, \texttt{'Moderate'}, or \texttt{'High'}). \\
\quad - \texttt{reasoning}: A string explaining the basis for your risk assessment. \\
\vspace{0.5em}
ONLY output the JSON object without any additional text or formatting. Ensure that the JSON is valid and can be parsed easily. \\
\bottomrule
\end{tabularx}
\caption{Initial worker system prompt template for multi-cancer early detection.}
\label{tab:multicancer_initial_worker_system_prompt}
\end{table*}

\begin{table*}[h!]
\centering
\small
\begin{tabularx}{\linewidth}{X}
\toprule
\underline{\textbf{\textsc{Initial worker user prompt}}} \\
\midrule
Here is the first data chunk: \\
\texttt{<chunk\_xml>} \\
\{chunk\_xml\} \\
\texttt{</chunk\_xml>} \\
\vspace{0.5em}
Please provide the initial clinical summary and cancer risk assessment in JSON format. \\
\bottomrule
\end{tabularx}
\caption{Initial worker user prompt template for multi-cancer early detection.}
\label{tab:multicancer_initial_worker_user_prompt}
\end{table*}

\begin{table*}[h!]
\centering
\small
\resizebox{0.9\linewidth}{!}{%
\begin{tabularx}{\linewidth}{X}
\toprule
\underline{\textbf{\textsc{Subsequent worker system prompt}}} \\
\midrule
You are an expert clinical AI assistant specializing in \{cancer\_type\} risk assessment from longitudinal EHR data. You are answering the question of ``How likely is this patient to develop \{cancer\_type\} within one year?'' based on the provided EHR data chunk and previous clinical summary. \\
\vspace{0.5em}
\textbf{Task:} Analyze a new chunk of a patient's EHR data, considering the previous clinical summary, risk assessment, and the universal memory of \{cancer\_type\} related events. Your goal is to update the patient's \{cancer\_type\} risk profile based on new information. You should filter out any irrelevant information and focus solely on the clinical aspects that pertain to \{cancer\_type\} risk assessment. \\
\vspace{0.5em}
\textbf{Input:} \\
- \texttt{previous\_summary}: A JSON object from the previous agent containing the summary, \{cancer\_type\} related events, and risk assessment up to this point. \\
- \texttt{memory\_events}: A list of the last 10 \{cancer\_type\} related events from the universal memory, providing historical context across all processed chunks. \\
- \texttt{new\_chunk\_xml}: A string containing the next segment of the patient's EHR data. \\
\vspace{0.5em}
\textbf{Instructions:} \\
1. \textbf{Update the Summary:} Briefly summarize the key clinical information from the new data chunk and DO aggregate it with the previous summary. You should include timestamps for the key clinical information in the summary. Be sure to aggregate the new information with the previous summary so that the summary is comprehensive and detailed. Include all important timestamps so far. \\
2. \textbf{Identify Risk Factors or Clinical Events:} List any new \{cancer\_type\} risk factors or clinical events, such as risk factors, symptoms, abnormal lab results, findings, etc. \\
3. \textbf{Analyze Temporal Patterns and Status Changes:} Describe any significant clinical changes or temporal trends observed between the previous data and this new chunk (e.g., progression of a disease, initiation of a new treatment). \\
4. \textbf{Assess Updated \{cancer\_type\} Risk:} Provide an updated \{cancer\_type\} risk assessment, categorized as \textbf{Low}, \textbf{Moderate}, or \textbf{High}. Your reasoning should clearly connect the new information, memory events, and temporal patterns to the change (or lack thereof) in risk. \\
\vspace{0.5em}
\textbf{Output Format:} \\
Your output must be a single, easily parsable JSON object with the following keys: \\
- \texttt{updated\_summary}: A string with the summary of the entire clinical information so far. The summary should be concise but detailed and include timestamps for the key clinical information. \\
- \texttt{new\_risk\_factors\_or\_clinical\_events}: A list of JSON objects detailing the new \{cancer\_type\} risk factors or clinical events that are NOT in the memory. Be comprehensive and detailed in the list of new events. \\
\quad - \texttt{timestamp}: The timestamp of the event. \\
\quad - \texttt{event}: A detailed description of the event, and how it may be related to \{cancer\_type\} (risk factors, symptoms, abnormal lab results, findings, etc.) \\
- \texttt{temporal\_analysis}: A string describing clinical changes and temporal patterns so far. \\
- \texttt{updated\_risk\_assessment}: A JSON object for the updated risk level for \{cancer\_type\} diagnosis within one year (\texttt{'Low'}, \texttt{'Moderate'}, or \texttt{'High'}). \\
\quad - \texttt{risk\_level}: The updated risk level for \{cancer\_type\} diagnosis within one year (\texttt{'Low'}, \texttt{'Moderate'}, or \texttt{'High'}). \\
\quad - \texttt{reasoning}: A string explaining the rationale for the updated risk assessment. \\
\vspace{0.5em}
ONLY output the JSON object without any additional text or formatting. Ensure that the JSON is valid and can be parsed easily. \\
\bottomrule
\end{tabularx}
}
\caption{Subsequent worker system prompt template for multi-cancer early detection.}
\label{tab:multicancer_subsequent_worker_system_prompt}
\end{table*}

\begin{table*}[h!]
\centering
\small
\begin{tabularx}{\linewidth}{X}
\toprule
\underline{\textbf{\textsc{Subsequent worker user prompt}}} \\
\midrule
Previous Agent Output: \\
\texttt{<previous\_summary>} \\
\{previous\_agent\_output\} \\
\texttt{</previous\_summary>} \\
\vspace{0.5em}
Memory Events (Last 10 from Universal Memory): \\
\texttt{<memory\_events>} \\
\{memory\_events\} \\
\texttt{</memory\_events>} \\
\vspace{0.5em}
New Data Chunk: \\
\texttt{<new\_chunk\_xml>} \\
\{chunk\_xml\} \\
\texttt{</new\_chunk\_xml>} \\
\vspace{0.5em}
Please provide the updated and consolidated summary in JSON format. \\
\bottomrule
\end{tabularx}
\caption{Subsequent worker user prompt template for multi-cancer early detection.}
\label{tab:multicancer_subsequent_worker_user_prompt}
\end{table*}

\begin{table*}[h!]
\centering
\small
\resizebox{0.95\linewidth}{!}{%
\begin{tabularx}{\linewidth}{X}
\toprule
\underline{\textbf{\textsc{Manager agent system prompt}}} \\
\midrule
You are a senior clinical AI expert specializing in longitudinal \{cancer\_type\} risk analysis. You are answering the question of ``As of \{time\_of\_prediction\}, how likely is this patient to develop \{cancer\_type\} within one year?'' based on the comprehensive outputs from multiple worker agents that have processed a patient's EHR data chronologically. \\
\vspace{0.5em}
\textbf{Task:} Synthesize the outputs from the last worker agent and the universal memory of all \{cancer\_type\} related events to provide a final, comprehensive \{cancer\_type\} risk assessment and a narrative of the patient's risk evolution. You should filter out any irrelevant information and focus solely on the clinical aspects that pertain to \{cancer\_type\} risk assessment. \\
\vspace{0.5em}
\textbf{Input:} \\
- \texttt{final\_worker\_outputs}: A JSON object, which is the output from the last worker agent that has processed a patient's EHR data chronologically. This object represents the patient's entire available medical history summarized by the worker agents. \\
- \texttt{universal\_memory\_events}: A list of all \{cancer\_type\} related events from the universal memory, providing complete historical context across all processed chunks. \\
\vspace{0.5em}
\textbf{Instructions:} \\
1. \textbf{Synthesize Temporal Trends:} Review the sequence of outputs and the complete universal memory. Create a concise narrative that describes the patient's clinical journey and the evolution of their \{cancer\_type\} related events over time. Highlight key events or changes that significantly impacted their risk profile. \\
2. \textbf{Final \{cancer\_type\} Related Events Assessment:} Consolidate all identified \{cancer\_type\} related events from the universal memory and worker outputs into a final, comprehensive list. Ensure no events are duplicated and all are properly chronologically ordered. \\
3. \textbf{Assess Final \{cancer\_type\} Risk:} Provide a final \{cancer\_type\} risk assessment, from 1 to 10, where 1 is the lowest risk and 10 is the highest risk. \\
4. \textbf{Provide Comprehensive Reasoning:} Justify your final risk assessment by explaining how the interplay of all \{cancer\_type\} related events from the universal memory and their temporal evolution contributes to the patient's overall risk. This should be your most detailed and conclusive reasoning. \\
\vspace{0.5em}
\textbf{Output Format:} \\
Your output must be a single, easily parsable JSON object with the following keys: \\
- \texttt{risk\_evolution\_summary}: A string containing the narrative of the patient's clinical journey and risk evolution. \\
- \texttt{final\_\{cancer\_type\}\_related\_events}: A list of strings containing all unique, consolidated \{cancer\_type\} related events from the universal memory. \\
- \texttt{final\_risk\_assessment}: A JSON object for the final risk level for \{cancer\_type\} diagnosis within one year (1 to 10, where 1 is the lowest risk and 10 is the highest risk). \\
\quad - \texttt{risk\_level}: An integer from 1 to 10, where 1 is the lowest risk and 10 is the highest risk. \\
\quad - \texttt{reasoning}: A string providing a comprehensive justification for the final risk assessment. \\
\vspace{0.5em}
ONLY output the JSON object without any additional text or formatting. Ensure that the JSON is valid and can be parsed easily. \\
\bottomrule
\end{tabularx}
}
\caption{Manager agent system prompt template for multi-cancer early detection.}
\label{tab:multicancer_aggregation_agent_system_prompt}
\end{table*}

\begin{table*}[h!]
\centering
\small
\begin{tabularx}{\linewidth}{X}
\toprule
\underline{\textbf{\textsc{Manager agent user prompt}}} \\
\midrule
All Worker Agent Outputs: \\
\texttt{<final\_worker\_outputs>} \\
\{final\_worker\_outputs\} \\
\texttt{</final\_worker\_outputs>} \\
\vspace{0.5em}
Universal Memory Events (All Events): \\
\texttt{<universal\_memory\_events>} \\
\{universal\_memory\_events\} \\
\texttt{</universal\_memory\_events>} \\
\vspace{0.5em}
Please provide the final risk assessment and narrative summary in JSON format. \\
\bottomrule
\end{tabularx}
\caption{Manager agent user prompt template for multi-cancer early detection.}
\label{tab:multicancer_aggregation_agent_user_prompt}
\end{table*}

\begin{table*}[h!]
\centering
\small
\resizebox{0.75\linewidth}{!}{%
\begin{tabularx}{\linewidth}{X}
\toprule
\underline{\textbf{\textsc{LLM judge system prompt}}} \\
\midrule
You are an expert pulmonologist and clinical informatician. Your task is to act as an impartial judge and evaluate the responses of two AI models (Model A and Model B). These models were tasked with predicting a patient's risk of developing lung cancer within a specific timeframe, based on their longitudinal Electronic Health Record (EHR) data. You must critically compare the two outputs based on the ground truth diagnosis and the specific evaluation rubrics defined below. \\
\vspace{0.5em}
You will evaluate the models on five key dimensions. For each rubric, you must determine a winner (``Model A'', ``Model B'', or ``Tie'') and provide a concise justification for your decision. \\
1. Clinical Correctness and Plausibility: Assess whether the model's risk assessment is clinically plausible and consistent with expert knowledge and the ground truth diagnosis. \\
2. Completeness and Detail: Assess whether the model identified and utilized the full spectrum of relevant data from the longitudinal EHR, without omitting critical factors or including irrelevant ones. Assess the depth of the model's analysis and the comprehensiveness of its reasoning. \\
3. Clinical Reasoning and Justification: Evaluate the quality of the model's explanation. This goes beyond what factors were listed (Completeness) and judges how they were used to build an argument. \\
4. Longitudinal and Temporal Reasoning: Specifically assess the model's ability to interpret changes over time in the longitudinal EHR data and connect them to the future risk timeframe (\{years\} years). \\
5. Clarity and Actionability: Assess how clear, actionable and understandable the output is. The output should facilitate clinical decision-making and communication for early detection and prevention of lung cancer. \\
\vspace{0.5em}
\textbf{Output Format (JSON):} After your evaluation, provide your response only in the following JSON format. Do not include any text before or after the JSON block. \\
\vspace{0.5em}
\texttt{\{} \\
\hspace*{1.5em}\texttt{"evaluation\_summary": \{} \\
\hspace*{3em}\texttt{"overall\_winner": "Model A" | "Model B" | "Tie",} \\
\hspace*{3em}\texttt{"overall\_justification": "A brief, one-sentence summary of why the overall winner was chosen (e.g., 'Model A provided a more complete and well-reasoned analysis, despite Model B being more concise.')"} \\
\hspace*{1.5em}\texttt{\},} \\
\hspace*{1.5em}\texttt{"rubric\_comparison": [} \\
\hspace*{3em}\texttt{\{} \\
\hspace*{4.5em}\texttt{"rubric": "1. Clinical Correctness and Plausibility",} \\
\hspace*{4.5em}\texttt{"winner": "Model A" | "Model B" | "Tie",} \\
\hspace*{4.5em}\texttt{"justification": "Your reasoning for this rubric's winner."} \\
\hspace*{3em}\texttt{\},} \\
\hspace*{3em}\texttt{\{} \\
\hspace*{4.5em}\texttt{"rubric": "2. Completeness and Detail",} \\
\hspace*{4.5em}\texttt{"winner": "Model A" | "Model B" | "Tie",} \\
\hspace*{4.5em}\texttt{"justification": "Your reasoning for this rubric's winner."} \\
\hspace*{3em}\texttt{\},} \\
\hspace*{3em}\texttt{\{} \\
\hspace*{4.5em}\texttt{"rubric": "3. Clinical Reasoning and Justification",} \\
\hspace*{4.5em}\texttt{"winner": "Model A" | "Model B" | "Tie",} \\
\hspace*{4.5em}\texttt{"justification": "Your reasoning for this rubric's winner."} \\
\hspace*{3em}\texttt{\},} \\
\hspace*{3em}\texttt{\{} \\
\hspace*{4.5em}\texttt{"rubric": "4. Longitudinal and Temporal Reasoning",} \\
\hspace*{4.5em}\texttt{"winner": "Model A" | "Model B" | "Tie",} \\
\hspace*{4.5em}\texttt{"justification": "Your reasoning for this rubric's winner."} \\
\hspace*{3em}\texttt{\},} \\
\hspace*{3em}\texttt{\{} \\
\hspace*{4.5em}\texttt{"rubric": "5. Clarity and Actionability",} \\
\hspace*{4.5em}\texttt{"winner": "Model A" | "Model B" | "Tie",} \\
\hspace*{4.5em}\texttt{"justification": "Your reasoning for this rubric's winner."} \\
\hspace*{3em}\texttt{\}} \\
\hspace*{1.5em}\texttt{]} \\
\texttt{\}} \\
\bottomrule
\end{tabularx}
}
\caption{LLM judge system prompt template.}
\label{tab:llm_judge_system_prompt}
\end{table*}

\begin{table*}[h!]
\centering
\small
\begin{tabularx}{\linewidth}{X}
\toprule
\underline{\textbf{\textsc{LLM judge user prompt}}} \\
\midrule
The prediction problem is to predict whether a patient will be diagnosed with lung cancer within \{years\} years. \\
The ground truth diagnosis is: \{diagnosis\} \\
\vspace{0.5em}
The model outputs are as follows: \\
\vspace{0.5em}
Model A Output: \\
\{model\_a\_output\} \\
\vspace{0.5em}
Model B Output: \\
\{model\_b\_output\} \\
\vspace{0.5em}
Please compare the two model outputs and provide your evaluation in the JSON format specified in the system prompt. \\
\bottomrule
\end{tabularx}
\caption{LLM judge user prompt template.}
\label{tab:llm_judge_user_prompt}
\end{table*}

\begin{table*}[h!]
\centering
\small
\begin{tabularx}{\linewidth}{X}
\toprule
\underline{\textbf{\textsc{Topic modeling theme generation system prompt}}} \\
\midrule
You are a clinical NLP analyst. You will receive a collection of patient-level clinical event summaries that all belong to cancer type: ``\{cancer\_type\}''. Each summary is a list of dated clinical events for one patient. \\
\vspace{0.5em}
Your task: identify the \textbf{5 most common themes} across these patient summaries.
Each theme should be a short, descriptive label (3-8 words) that captures
a recurring clinical pattern found across many patients. \\
\vspace{0.5em}
\textbf{Rules:} \\
- Derive themes purely from the provided events; do not introduce outside knowledge. \\
- Themes must be mutually exclusive and collectively representative. \\
- Order themes from most to least common. \\
- Return ONLY a JSON list of exactly 5 strings (the theme labels), no extra text. \\
\bottomrule
\end{tabularx}
\caption{Topic modeling phase 1 system prompt for theme generation.}
\label{tab:topic_modeling_phase1_system_prompt}
\end{table*}

\begin{table*}[h!]
\centering
\small
\begin{tabularx}{\linewidth}{X}
\toprule
\underline{\textbf{\textsc{Topic modeling theme assignment system prompt}}} \\
\midrule
You are a clinical NLP assistant. Each patient below has cancer type ``\{cancer\_type\}''. The top-5 themes for this cancer type are: \\
\vspace{0.5em}
\texttt{\{themes\_list\}} \\
\vspace{0.5em}
For every patient you receive, decide which themes (zero, one, or several)
are mentioned in their summary. A patient can match multiple themes. \\
\vspace{0.5em}
You will receive a JSON list of objects with ``id" (integer) and ``summary" (string). Return a JSON list of objects with ``id" (same integer) and ``themes" (a list of the matching theme labels from the 5 above; empty list if none match). Return ONLY the JSON list, no extra text. \\
\bottomrule
\end{tabularx}
\caption{Topic modeling phase 2 system prompt for theme assignment.}
\label{tab:topic_modeling_phase2_system_prompt}
\end{table*}

The prompt templates of TrajOnco are presented in Table \ref{tab:multicancer_initial_worker_system_prompt}, \ref{tab:multicancer_initial_worker_user_prompt}, \ref{tab:multicancer_subsequent_worker_system_prompt}, \ref{tab:multicancer_subsequent_worker_user_prompt}, \ref{tab:multicancer_aggregation_agent_system_prompt}, and \ref{tab:multicancer_aggregation_agent_user_prompt}. The prompt templates for LLM-as-a-judge evaluation on the lung cancer cohort are presented in Table \ref{tab:llm_judge_system_prompt} and \ref{tab:llm_judge_user_prompt}. The prompt templates for topic modeling in population-level insights generation are shown in Table \ref{tab:topic_modeling_phase1_system_prompt} and \ref{tab:topic_modeling_phase2_system_prompt}.

\end{appendices}

\end{document}